\pdfoutput=1
\documentclass{article}
\usepackage{preprint,times}

\usepackage{amsmath,amsfonts,bm}

\def\eqref#1{equation~\ref{#1}}

\def\1{\bm{1}}

\def\rk{{\textnormal{k}}}

\DeclareMathAlphabet{\mathsfit}{\encodingdefault}{\sfdefault}{m}{sl}
\SetMathAlphabet{\mathsfit}{bold}{\encodingdefault}{\sfdefault}{bx}{n}

\newcommand{\R}{\mathbb{R}}

\usepackage{url}
\usepackage{graphicx}
\usepackage{booktabs}
\usepackage{multirow}
\usepackage{xcolor}
\usepackage{caption}
\usepackage{subcaption}
\usepackage{enumitem}
\usepackage{tabularx}
\usepackage{colortbl}
\usepackage{array}
\usepackage{amssymb}
\usepackage{placeins}
\usepackage{float}
\usepackage{tikz}
\usetikzlibrary{positioning,fit,calc,backgrounds,arrows.meta,shadows}

\definecolor{linkblue}{rgb}{0.21,0.49,0.74}
\usepackage[breaklinks,colorlinks,allcolors=linkblue]{hyperref}

\newcommand{\method}{Luce}
\newcommand{\ie}{i.e.\@}
\newcommand{\eg}{e.g.\@}

\newcommand{\bp}{\bm{p}}
\newcommand{\bs}{\bm{s}}
\newcommand{\bq}{\bm{q}}
\newcommand{\bc}{\bm{c}}
\newcommand{\bz}{\mathbf{z}}
\newcommand{\bF}{\mathbf{F}}
\newcommand{\bG}{\mathbf{G}}

\newcommand{\oursrow}{\rowcolor{gray!10}}

\makeatletter
\def\paragraph{\@startsection{paragraph}{4}{\z@}{0.2ex plus 0.2ex minus .05ex}{-1em}{\normalsize\bf}}
\makeatother

\title{
Luce: Relightable Gaussians for 3D Asset Generation
}

\author{\hbox to 0.8\linewidth{\bf Mayank Singh\hfil Michele Stoppa\hfil Alvise Memo\hfil Rui Yu}\\
\hbox to 0.8\linewidth{\bf Harsha Kalli\hfil Srimanth Gunturi\hfil Muhammad Ahmed Riaz}\\
\hbox spread 30pt{\bf Behrooz Shahsavari\hfil Waleed Abdulla\hfil David E. Jacobs}\\
\mdseries Apple
}

\preprintfinalcopy
\begin{document}

\maketitle
\lhead{}

\begin{figure}[H]
\begin{center}
\begin{tikzpicture}[inner sep=0pt, outer sep=0pt]
\node[anchor=north west] (strip-top) at (0,0)
     {\includegraphics[width=0.86\linewidth,trim=1024 0 120 0,clip]
        {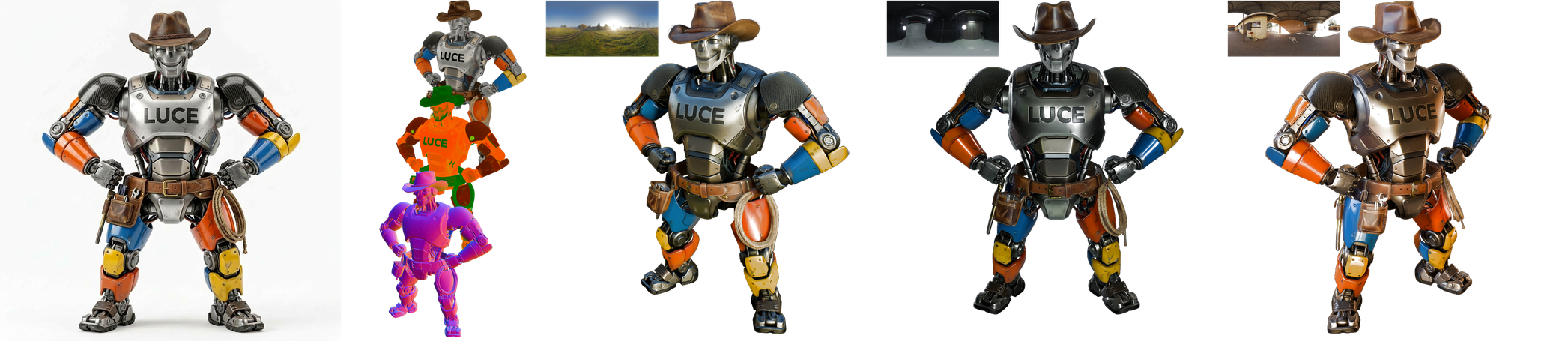}};
\node[anchor=east, draw=gray, line width=0.25pt, xshift=-2pt] at (strip-top.west)
     {\includegraphics[width=0.14\linewidth,trim=120 0 3806 0,clip]
        {figures/banner/luce_robot_snapshot_demo_seed_14_r145_cz001_bright.png}};

\node[anchor=north west] (strip-bot) at (strip-top.south west)
     {\includegraphics[width=0.86\linewidth,trim=1024 0 64 0,clip]
        {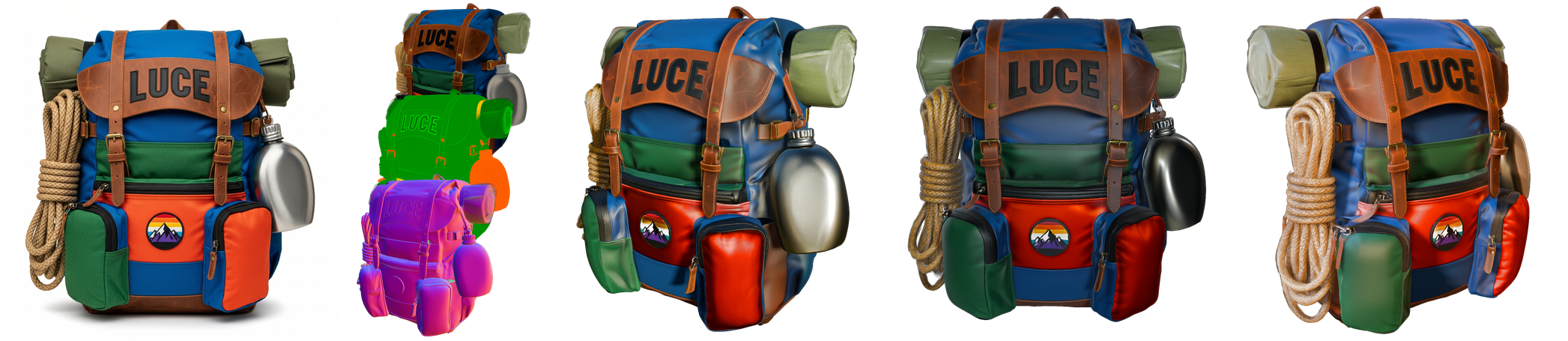}};
\node[anchor=east, draw=gray, line width=0.25pt, xshift=-2pt, inner sep=0pt,
      minimum width=0.14\linewidth, minimum height=0.1828\linewidth]
     at (strip-bot.west)
     {\includegraphics[width=0.14\linewidth,trim=40 0 3740 0,clip]
        {figures/banner/luce_backpack_snapshot_demo_seed_15_r160_p15_cz-005_no_ibl_bright.png}};

\end{tikzpicture}
\end{center}
\vspace{-0.3cm}
\caption{Our method \textbf{\method{}} generates relightable 3D assets from a single image.
Each asset consists of Gaussians encoding three modalities of a physically based rendering (PBR) material: albedo, metallic-roughness, and surface normals.
These modalities enable standard PBR shading, rendering the Gaussians under novel illumination.
From left to right, the columns show: the input condition image; the generated PBR modalities stacked vertically---albedo (top), metallic-roughness (middle, metallic in Red, roughness in Green channel), and surface normals (bottom); and the generated 3D asset rendered under three environment maps, previewed as small strips above each render (top row).
}
\label{fig:teaser}
\end{figure}

\begin{abstract}
High-fidelity image-to-3D generation requires a 3D representation that captures both geometry and appearance. However, preserving fine detail across the physically based rendering (PBR) modalities needed for relighting remains challenging.
To address this, we propose \method{}, a 3D representation that unifies geometry and PBR materials within a voxelized multimodal Gaussian cloud, using dedicated Gaussian primitives for albedo, metallic-roughness, and surface normals.
A variational autoencoder compresses this representation into a unified material-aware latent space.
A rectified-flow transformer generates this latent from a single image using multi-layer features from a pretrained image encoder that preserve both semantic context and fine spatial detail. The latent is then decoded into relightable PBR Gaussians and an optional textured mesh with a tangent-space normal map.
On Toys4K, \method{} achieves state-of-the-art single-image-to-3D generation, improving FID by 28\% over the strongest baseline. We further evaluate \method{} on a benchmark of AI-generated images depicting diverse subjects and materials, where it improves the CLIP image-alignment score over the best baseline (0.8519 vs.\ 0.8299).
\method{} generates relightable, geometrically accurate, and materially faithful assets that preserve fine details such as text, logos, and inscriptions.
\end{abstract}

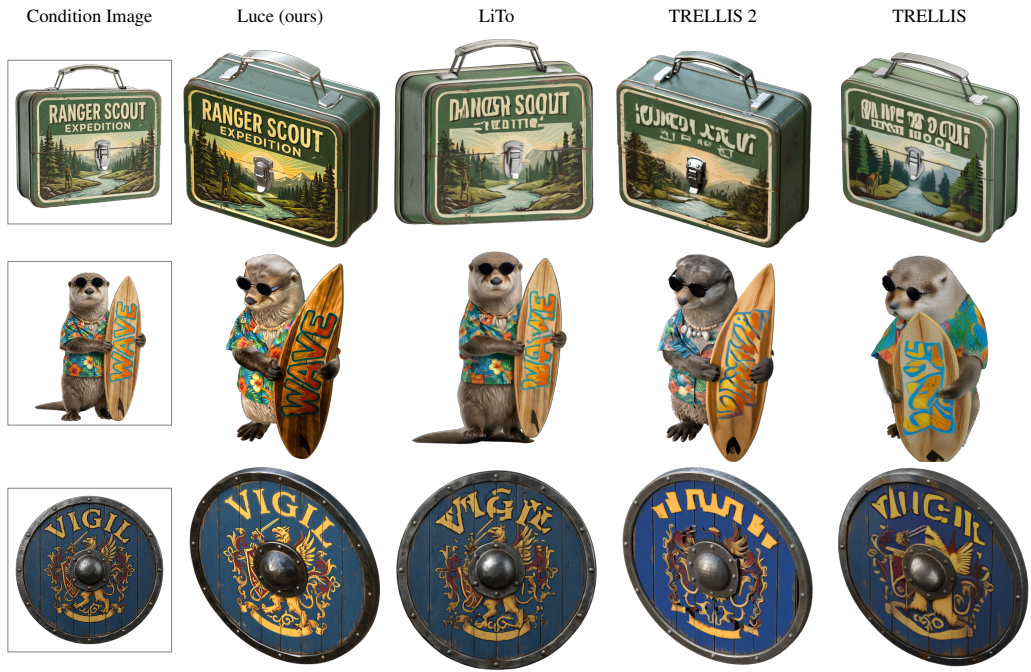
\begin{figure}[t]
\centering

\newcommand{\luceCmpRoot}{figures/gen_compare/per_method}
\newcommand{\luceCmpW}{0.205\linewidth}
\newcommand{\luceCondW}{0.155\linewidth}

\newcommand{\luceCondImgTC}[3]{%
\begin{tikzpicture}[baseline=(current bounding box.center), inner sep=0pt, outer sep=0pt]
  \node[draw=gray, line width=0.25pt, minimum width=\luceCondW, minimum height=\luceCondW, inner sep=0pt, path picture={
    \node at (path picture bounding box.center)
         {\includegraphics[width=#3,trim=#2,clip]{\luceCmpRoot/#1/cond_image.png}};
  }] {};
\end{tikzpicture}%
}

\newcommand{\luceMethodImgT}[4][0000]{%
\begin{tikzpicture}[baseline=(current bounding box.center), inner sep=0pt, outer sep=0pt]
  \node {\includegraphics[width=\luceCmpW,trim=#4,clip]{\luceCmpRoot/#2/#3_color.#1.png}};
\end{tikzpicture}%
}

\newcommand{\luceMethodImgTW}[5][0000]{%
\begin{tikzpicture}[baseline=(current bounding box.center), inner sep=0pt, outer sep=0pt]
  \node {\includegraphics[width=#5,trim=#4,clip]{\luceCmpRoot/#2/#3_color.#1.png}};
\end{tikzpicture}%
}

\newcommand{\luceCmpRowTC}[5][0000]{%
\luceCondImgTC{#2}{#4}{#5} &
\luceMethodImgT{#2}{luce_gs}{#3} &
\luceMethodImgT{#2}{lito}{#3} &
\luceMethodImgT{#2}{trellis2}{#3} &
\luceMethodImgT[#1]{#2}{trellis1_gs}{#3} \\
}

\setlength{\tabcolsep}{0pt}
\renewcommand{\arraystretch}{0.70}
\begin{tabular}{@{}ccccc@{}}
{\scriptsize Condition Image} &
{\scriptsize \method{}~(ours)} &
{\scriptsize LiTo} &
{\scriptsize TRELLIS~2} &
{\scriptsize TRELLIS} \\

\luceCondImgTC{image_043}{70 20 70 70}{0.180\linewidth} &
\luceMethodImgTW{image_043}{luce_gs}{70 20 70 70}{0.190\linewidth} &
\luceMethodImgT{image_043}{lito}{70 20 70 70} &
\luceMethodImgTW{image_043}{trellis2}{70 20 70 70}{0.190\linewidth} &
\luceMethodImgTW[0001]{image_043}{trellis1_gs}{70 20 70 70}{0.180\linewidth} \\[-13pt]
\luceCondImgTC{image_136}{100 50 100 50}{0.130\linewidth} &
\luceMethodImgT{image_136}{luce_gs}{100 50 100 50} &
\luceMethodImgTW{image_136}{lito}{100 50 100 50}{0.185\linewidth} &
\luceMethodImgT{image_136}{trellis2}{100 50 100 50} &
\luceMethodImgTW{image_136}{trellis1_gs}{100 50 100 50}{0.200\linewidth} \\[-6pt]
\luceCondImgTC{image_065}{0 0 0 0}{0.150\linewidth} &
\luceMethodImgT{image_065}{luce_gs}{100 50 100 50} &
\luceMethodImgTW{image_065}{lito}{100 50 100 50}{0.190\linewidth} &
\luceMethodImgT{image_065}{trellis2}{100 50 100 50} &
\luceMethodImgT{image_065}{trellis1_gs}{100 50 100 50} \\

\end{tabular}

\caption{\textbf{Legible text on generated 3D assets.}
\method{} compared with LiTo~\citep{chang2026lito}, TRELLIS~2~\citep{xiang2025trellis2}, and TRELLIS~\citep{xiang2024trellis} on single-image-to-3D generation; each row shows the input condition image and one rendered view of each method's generated 3D asset. LiTo generates assets aligned to the input view, whereas other methods generate in a canonical orientation. \method{} keeps surface text and markings legible where baselines distort them.}
\label{fig:gen_compare_luce}
\end{figure}

\section{Introduction}
\label{sec:intro}

Diffusion and flow-matching models have largely closed the realism gap in 2D image generation~\citep{rombach2022stable,flux2024}, where images share one representation: a dense pixel grid.
3D has many, spanning implicit fields and explicit primitives~\citep{park2019deepsdf,mildenhall2020nerf,shen2023flexicubes,kerbl20233dgs}, and that choice sets a ceiling on what a generative model can produce.
To render and relight assets in standard pipelines, the representation must support fine geometric detail and physically based materials together.

We introduce \method{} (Fig.~\ref{fig:teaser}), a representation that meets these requirements: it unifies fine geometry and physically based materials in a single relightable form.
In \method{}, an asset is a collection of Gaussians attached to a sparse set of voxels that intersect the object's surface.
Within each voxel, we encode a set of Gaussians for each PBR modality: albedo, metallic-roughness, and surface normal direction.
This representation is a complete PBR material description in a 3D-native format that drops directly into standard rendering pipelines.
It also compresses into a compact, diffusible latent.
The per-modality encoding of geometry and appearance helps to preserve high-frequency details that prior approaches often smooth out, including legible text on generated 3D surfaces, a setting where existing image-to-3D methods typically struggle.
This observation is supported by our quantitative studies, where \method{} achieves the lowest FID on Toys4K (20.99), improving on the strongest baselines, TRELLIS~2~\citep{xiang2025trellis2} (29.22) and LiTo (29.76), by more than 8 FID.

Our contributions are:

\textbf{(i)~A multimodal PBR Gaussian representation} (Sec.~\ref{sec:representation}). Our representation jointly models geometry and appearance, parameterizing albedo, metallic-roughness, and surface normals directly on Gaussian primitives. It supports relighting under image-based illumination via deferred shading in a standard PBR reflectance model.

\textbf{(ii)~A unified latent for joint geometry and material generation} (Sec.~\ref{sec:vae}). We learn a compact latent over the PBR Gaussian representation that encodes geometry and all material modalities together, so a rectified-flow transformer generates a complete relightable asset from a single image. The latent decodes directly into relightable PBR Gaussians for rendering; optionally, we extract a mesh and bake the decoded Gaussians into its PBR texture maps, yielding a textured mesh.

\textbf{(iii)~Multi-layer image conditioning} (Sec.~\ref{sec:flow}). We propose combining DINOv2~\citep{oquab2024dinov2} features from shallow and deep encoder layers, preserving fine spatial detail (\eg{}, logos, labels, and inscriptions on the asset) critical for high-fidelity 3D generation (see Fig.~\ref{fig:gen_compare_luce} and Fig.~\ref{fig:multi_layer_dino_ablation}).

\textbf{(iv)~Tangent-space normal map transfer} (Sec.~\ref{sec:normal_transfer}). \method{}'s normal Gaussians learn fine surface detail (\eg{}, engravings, fabric weave, embossed text) from the asset's authored normal maps, beyond what mesh geometry encodes. At inference, we bake these decoded normals onto the extracted mesh as tangent-space normal maps, adding high-frequency detail at zero polygon-count cost.

\section{Related Work}
\label{sec:related}

\paragraph{3D representations for generation.}
A generative model's representation space largely determines what it can faithfully capture.
Implicit level sets~\citep{mescheder2019occupancy} admit arbitrary genus without a template, but only as closed, watertight surfaces.
More recent implicit approaches scale to higher fidelity but exclusively target geometry: TripoSG~\citep{li2024triposg} via SDFs and Direct3D-S2~\citep{wu2024direct3d} via sparse-voxel fields; both inherit the same watertight requirement.
3D Gaussian Splatting (3DGS)~\citep{kerbl20233dgs} renders appearance explicitly and cheaply, and a series of feed-forward and reconstruction techniques~\citep{tang2024dreamgaussian,tang2024lgm,zhang2024gslrm,szymanowicz2024splatter,zhang2024gaussiancube} extends it to image- or text-to-3D. These approaches do not model materials; like NeRF~\citep{mildenhall2020nerf}, they bake the lighting environment into the representation.
Structured-latent methods take a different route, combining sparse voxels with learned per-voxel features: TRELLIS~\citep{xiang2024trellis} stores DINOv2 features~\citep{oquab2024dinov2} and decodes to multiple formats, TRELLIS~2~\citep{xiang2025trellis2} replaces these with native O-Voxel geometry and per-voxel PBR attributes, and LiTo~\citep{chang2026lito} tokenizes surface light fields via Perceiver~IO~\citep{jaegle2022perceiverio}.
\method{} pairs the rendering efficiency of 3DGS with per-modality PBR decomposition in a structured latent, without mesh dependence or baked lighting.

\paragraph{3D generative models.}
Latent-space approaches have become the dominant paradigm for 3D generation.
Early methods such as Shap-E~\citep{jun2023shape} and 3DTopia-XL~\citep{chen2024primx} generate implicit-function parameters or primitive-based latents; recent ones build on Diffusion Transformers~(DiT)~\citep{peebles2023dit} and flow matching~\citep{lipman2023flow,liu2023rectifiedflow}.
Among DiT-based approaches, TRELLIS~\citep{xiang2024trellis} and TRELLIS~2~\citep{xiang2025trellis2} use a structure-then-latent paradigm, LiTo~\citep{chang2026lito} conditions a single DiT on DINOv2~\citep{oquab2024dinov2}, and Step1X-3D~\citep{step1x3d2025} and Hunyuan3D~2.1~\citep{hunyuan3d2025} pair a DiT shape generator with a separate multi-view texture-diffusion stage.
\method{} also uses rectified-flow transformers, but operates in a unified PBR Gaussian latent so geometry and materials are generated jointly.

\paragraph{PBR and relightable 3D.}
Relighting requires decomposing appearance into material properties.
Per-scene inverse rendering does this for a single asset~\citep{munkberg2022nvdiffrec,zhang2021nerfactor,liang2024gsir,jiang2024gaussianshader,gao2023relightable3dgs}, but does not generalize to novel objects.
The GS-based ones among them tie per-Gaussian normals to the reconstructed surface, via a flattened Gaussian's shortest axis or depth-derived pseudo-normals. \method{} instead predicts normals as a separate modality, so they carry detail finer than the geometry.
Recent methods attach PBR attributes to Gaussians in feed-forward or generative settings, each with a constraint: TexGaussian~\citep{xiong2025texgaussian} requires an input mesh; RelitLRM~\citep{zhang2025relitlrm} models appearance with spherical harmonics rather than explicit PBR; MGM~\citep{ye2025mgm} is text-conditioned and runs in multiple stages; MatSpray~\citep{langsteiner2026matspray} optimizes per-scene from multiple views.
Among generative models, TRELLIS~2~\citep{xiang2025trellis2} produces PBR-complete assets via a mesh-centric representation, and LiTo~\citep{chang2026lito} captures view-dependent effects through spherical harmonics that cannot be decomposed into materials.
A separate line of work estimates normals from images~\citep{bae2024dsine,ye2024stablenormal}; we render decoded normal Gaussians and bake them directly as tangent-space maps (Sec.~\ref{sec:normal_transfer}).
\method{} learns a compact diffusible latent from large-scale PBR data that jointly generates all material modalities and decodes to PBR Gaussians and textured meshes.

\section{Method}
\label{sec:method}

\begin{figure}[t]
\begin{center}
\includegraphics[width=\linewidth]{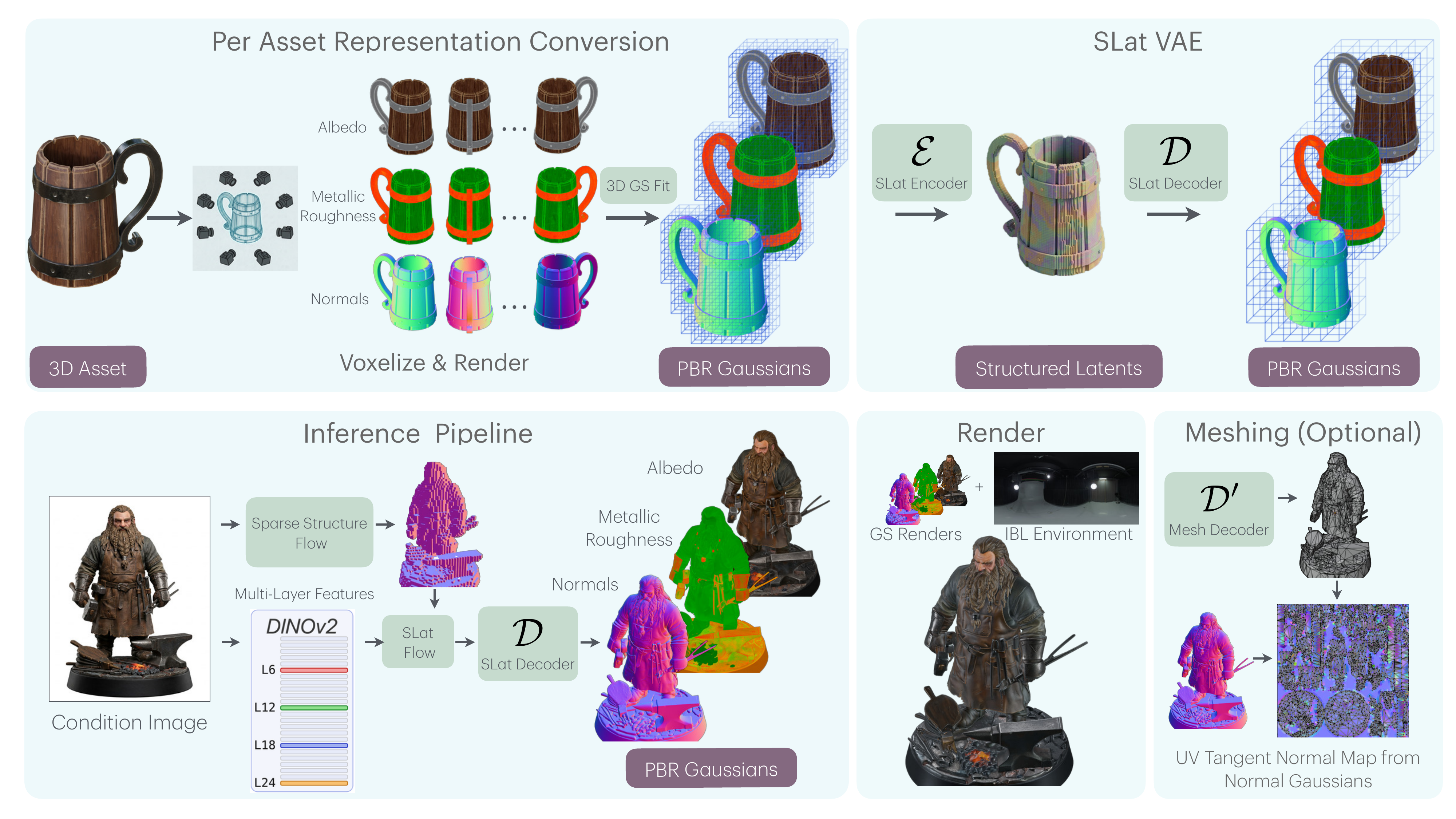}
\end{center}
\vspace{-0.3cm}
\caption{\textbf{Overview of \method{}.}
\textbf{(Top)~Representation and SLatVAE.} Given a 3D PBR asset, here a Toys4K~\citep{stojanov2021toys4k} sample, we render multiview images and fit per-modality Gaussian splats (albedo, metallic-roughness, normals) on a sparse voxel grid. A structured latent VAE (SLatVAE) encodes this representation into a compact, diffusible latent and decodes it back to PBR Gaussians.
\textbf{(Bottom)~Generation pipeline.} Given a single input image, a sparse-structure flow first predicts the sparse voxel layout. Conditioned on multi-layer DINOv2 features, SLatFlow then generates a PBR Gaussian latent at each active voxel. The structured latent decodes into relightable PBR Gaussians and also serves as input to the mesh decoder, which yields textured meshes with tangent-space normal maps and a complete PBR material set.}
\label{fig:overview}
\end{figure}

As shown in Fig.~\ref{fig:overview}, \method{} represents assets as multimodal PBR Gaussian clouds, compresses this representation into a compact latent with a structured latent VAE (SLatVAE), and trains an image-conditioned rectified-flow model, the structured latent flow (SLatFlow), to generate the latent.

\subsection{Multimodal PBR Gaussian Representation}
\label{sec:representation}

The standard PBR pipeline models surface appearance as a function of independent modalities: diffuse reflectance (albedo), material properties (metallic, roughness), and surface orientation (normals).
We mirror this decomposition in 3D by giving each occupied voxel three dedicated Gaussian sets: albedo $\bG^{\text{alb}}$ (diffuse surface color), metallic-roughness $\bG^{\text{mr}}$ (combined metallic and roughness), and normal $\bG^{\text{nor}}$ (per-point surface orientation).
We denote this set of modalities as $\mathcal{M} = \{\text{alb},\,\text{mr},\,\text{nor}\}$.
Rather than extending a single set of Gaussians with PBR attributes, we keep the three Gaussian sets geometrically independent: each has its own positions, scales, rotations, and opacities within the shared voxel structure.
This keeps the pipeline simple, since each modality is fit and splatted with standard 3DGS against its own intrinsic images, and it lets each modality place its Gaussians where its own detail is richest.
Because the normal modality is stored explicitly rather than derived from the surface shape, its normals can deviate from the underlying surface geometry, a property we leverage in Sec.~\ref{sec:normal_transfer} for tangent-space normal map transfer.

\paragraph{Rendering PBR Gaussians.}
Each modality is splatted independently via standard 3DGS alpha-compositing~\citep{kerbl20233dgs}, producing per-pixel albedo $\bm a$, metallic $\kappa$, roughness $\alpha$, surface normal $\bm n$, and opacity $o$. Normals are renormalized to unit length after compositing.
We shade these splatted PBR buffers using split-sum image-based lighting~\citep{karis2013real} with the Cook--Torrance microfacet BRDF~\citep{cook1982reflectance}:
\begin{equation}
L_o^{\mathrm{IBL}}(\bm{\omega}_o)
\approx
\frac{(1-\kappa)\,\bm a}{\pi}\,E(\bm n)
\;+\;
L_{\mathrm{pref}}(\bm r,\alpha)\,
\left[\,
F_0\,A(\mu_o,\alpha) + B(\mu_o,\alpha)
\right],
\label{eq:pbr_shading}
\end{equation}
where $\bm{\omega}_o$ is the outgoing/view direction toward the camera, $\bm r = 2(\bm n\cdot\bm{\omega}_o)\,\bm n - \bm{\omega}_o$, $\mu_o = \operatorname{clamp}(\bm n\cdot\bm{\omega}_o,\,0,\,1)$, and $F_0 = (1-\kappa)\,0.04 + \kappa\,\bm a$.
Here $E(\bm n)$ is the diffuse irradiance map, $L_{\mathrm{pref}}(\bm r,\alpha)$ is the GGX-prefiltered environment map sampled along the mirror reflection direction $\bm r$, and $A,B$ are the two channels of the split-sum BRDF lookup table indexed by normal--view cosine $\mu_o$ and roughness $\alpha$. The Fresnel base reflectance $F_0$ follows the standard metallic workflow: dielectrics use $0.04$, while metals take their colored specular reflectance from the albedo $\bm a$. This shading is evaluated on the splatted Gaussian outputs, with no mesh extraction or UV unwrapping required; see~\citet{karis2013real,lagarde2014moving} for the derivation from the rendering equation. For a qualitative comparison of this deferred renderer against Blender~EEVEE~\citep{Blender}, see Fig.~\ref{fig:vae_relight}.

Each Gaussian uses the standard 3DGS parameterization (position offset $\bp$ from the voxel center, anisotropic scale $\bs$, rotation quaternion $\bq$, opacity $o$) together with a single view-independent modality value $\bc$ (RGB for albedo, metallic and roughness for metallic-roughness, normal direction for normals). Unlike the original 3DGS, we store no spherical harmonics; view-dependent shading is produced analytically through PBR compositing (Eq.~\ref{eq:pbr_shading}).
With $i$ indexing voxels and $j$ indexing the $K$ Gaussians per modality, the complete per-voxel representation at voxel $i$ is:
\begin{equation}
    \bF_i = \bigl\{\bG_i^{m}\bigr\}_{m \in \mathcal{M}}, \quad
    \bG_i^{m} = \bigl\{(\bp_{ij}^{m},\, \bs_{ij}^{m},\, \bq_{ij}^{m},\, o_{ij}^{m},\, \bc_{ij}^{m})\bigr\}_{j=1}^{K},
    \label{eq:voxel_repr}
\end{equation}
where $K$ is the number of Gaussians per modality per voxel, $d$ is the voxel-grid resolution, and $N_v^{d^3}$ is the number of active voxels in the $d^3$ grid. The per-voxel bundles form $\mathcal{V} = \{\bF_i\}_{i=1}^{N_v^{d^3}}$, the asset's complete voxelized multimodal Gaussian representation.
We store $\bs$ in log-space, $\bq$ as a unit quaternion, and pass $o$ through a sigmoid.
We build each asset's $\mathcal{V}$ in a preprocessing step, fitting the $K$ Gaussians of each modality independently against multi-view PBR renders of the asset; these fitted representations serve as input to the SLatVAE.

\subsection{Variational Autoencoder}
\label{sec:vae}

The raw multimodal Gaussian cloud $\mathcal{V}$ is high-dimensional and sparsely structured, so modeling it directly with a generative model is computationally expensive.
We compress it into a compact per-voxel latent with the SLatVAE (architecture in Appendix~\ref{app:hyperparams} and Fig.~\ref{fig:vae_arch}).
The key design choice is to \emph{trade spatial resolution for per-voxel density}: the encoder downsamples the voxel grid, and the decoder compensates by predicting more Gaussians per voxel. This keeps the latent small enough to diffuse cheaply while still reconstructing fine material and geometric detail.

\paragraph{Encoder.}
$\mathcal{E}$ takes the concatenation of all three PBR Gaussian sets per voxel (the standard GS parameters across $|\mathcal{M}|=3$ modalities, with $K$ Gaussians per modality) and is implemented as a sparse transformer with shifted-window attention~\citep{liu2021swin} that downsamples the input grid from resolution $d$ to a coarser $d' < d$, producing a compact latent $\bz = \mathcal{E}(\mathcal{V}) \in \R^{N_v^{{d'}^3} \times C}$, where $N_v^{{d'}^3}$ is the number of active voxels at the resolution $d'$ and $C$ is the latent channel dimension.

\paragraph{Decoder.}
$\mathcal{D}$ runs at the coarser latent resolution for efficiency and offsets the lost resolution by predicting more Gaussians per voxel per modality, recovering fine surface detail.
We train the SLatVAE end-to-end with a rendering-based reconstruction loss, as in TRELLIS~\citep{xiang2024trellis}: the decoded Gaussians are differentiably rendered per modality against the ground-truth intrinsic images, with a KL prior on the latent and light regularizers on Gaussian scale and opacity. The full objective and coefficients are in Appendix~\ref{app:hyperparams}.

\subsection{Flow-Based Generation}
\label{sec:flow}

We use rectified-flow transformers to generate in the SLatVAE's latent space, conditioned on a single image (bottom panel of Fig.~\ref{fig:overview}).
Generation proceeds in two stages: we first generate the object's sparse voxel structure (which voxels it intersects), then the latent features at each occupied voxel. We adopt this structure-then-latent approach from TRELLIS~\citep{xiang2024trellis}.

\paragraph{Sparse structure generation.}
We use the pretrained sparse-structure VAE and sparse-structure flow from TRELLIS~2~\citep{xiang2025trellis2} without modification.
Given a condition image, these models output the $N_v^{{d'}^3}$ active voxels that serve as the scaffold for the next stage.

\paragraph{Latent generation.}
SLatFlow then generates a PBR Gaussian latent at each active voxel, producing geometry and materials jointly.
It is a DiT-style transformer~\citep{peebles2023dit} adapted for sparse 3D latents: each sample's active voxels form a variable-length token sequence.
The diffusion timestep modulates each block via adaptive layer normalization; image features are injected through cross-attention.
Architectural details are in Appendix~\ref{app:hyperparams}.

\begin{figure}[t]
\centering

\newcommand{\mldRoot}{figures/snapshot_08arql_yaw180_r160_v35}
\newcommand{\mldCbox}{0.118\linewidth}
\newcommand{\mldExGap}{0.010\linewidth}
\newcommand{\mldMidGap}{0.020\linewidth}
\newcommand{\mldCondGap}{0.020\linewidth}%

\newcommand{\mldCondBox}[3]{%
\begin{tikzpicture}[baseline=(current bounding box.center), inner sep=0pt, outer sep=0pt]
  \node[draw=gray, line width=0.25pt, minimum width=\mldCbox, minimum height=\mldCbox, inner sep=0pt, path picture={
    \node at (path picture bounding box.center)
         {\includegraphics[width=#3,trim=#2,clip]{\mldRoot/#1_gen_luce_gs/cond_image.png}};
  }] {};
\end{tikzpicture}%
}

\newcommand{\mldModCascade}[6]{%
\begin{tikzpicture}[baseline=(current bounding box.center), inner sep=0pt, outer sep=0pt]
  \node[anchor=north west] at (2*#6,0)
       {\includegraphics[width=#4,trim=#3,clip]{\mldRoot/#1_gen_#2/albedo.0003.png}};
  \node[anchor=north west] at (#6,-#5)
       {\includegraphics[width=#4,trim=#3,clip]{\mldRoot/#1_gen_#2/mra.0003.png}};
  \node[anchor=north west] at (0,-2*#5)
       {\includegraphics[width=#4,trim=#3,clip]{\mldRoot/#1_gen_#2/normal.0003.png}};
\end{tikzpicture}%
}

\newcommand{\mldMethodCell}[8]{%
\begin{tikzpicture}[baseline=(current bounding box.center), inner sep=0pt, outer sep=0pt]
  \node[anchor=west, inner sep=0pt] (col) at (0,0)
       {\includegraphics[width=#4,trim=#3,clip]{\mldRoot/#1_gen_#2/color.0000_bright.png}};
  \node[anchor=west, inner sep=0pt] at ($(col.east)+(#8,0)$)
       {\mldModCascade{#1}{#2}{#3}{#5}{#6}{#7}};
\end{tikzpicture}%
}

\newcommand{\mldCellA}[1]{\mldMethodCell{image_150}{#1}{300 60 300 40}{0.076\linewidth}{0.032\linewidth}{0.033\linewidth}{0.007\linewidth}{0.004\linewidth}}
\newcommand{\mldCellB}[1]{\mldMethodCell{image_023}{#1}{150 55 78 28}{0.140\linewidth}{0.058\linewidth}{0.032\linewidth}{0.002\linewidth}{0.000\linewidth}}

\setlength{\tabcolsep}{0.5pt}
\renewcommand{\arraystretch}{1.0}
\begin{tabular}{@{}c@{\hskip\mldCondGap}c@{\hskip\mldMidGap}c@{\hskip\mldExGap}c@{\hskip\mldCondGap}c@{\hskip\mldMidGap}c@{}}
{\scriptsize\shortstack{Condition\\Image}} &
{\scriptsize\shortstack{\method{} (ours)\\Multi-layer}} &
{\scriptsize\shortstack{\method{} (ours)\\Single-layer}} &
{\scriptsize\shortstack{Condition\\Image}} &
{\scriptsize\shortstack{\method{} (ours)\\Multi-layer}} &
{\scriptsize\shortstack{\method{} (ours)\\Single-layer}} \\
\mldCondBox{image_150}{300 60 300 40}{0.052\linewidth} &
\mldCellA{luce_gs} &
\mldCellA{luce_single_gs} &
\mldCondBox{image_023}{150 55 78 28}{0.096\linewidth} &
\mldCellB{luce_gs} &
\mldCellB{luce_single_gs} \\
\end{tabular}

\caption{\textbf{Effect of multi-layer DINOv2 conditioning.} We compare \method{} trained with multi-layer DINOv2 features (layers 6, 12, 18, 24) against single-layer DINOv2 features (layer 24 only). Multi-layer conditioning preserves fine spatial detail from the condition image, including legible text and logos on the generated 3D surface. For each variant, the larger shaded render is shown next to a cascade of the per-modality decomposition (albedo, metallic-roughness, surface normals).}
\label{fig:multi_layer_dino_ablation}
\end{figure}
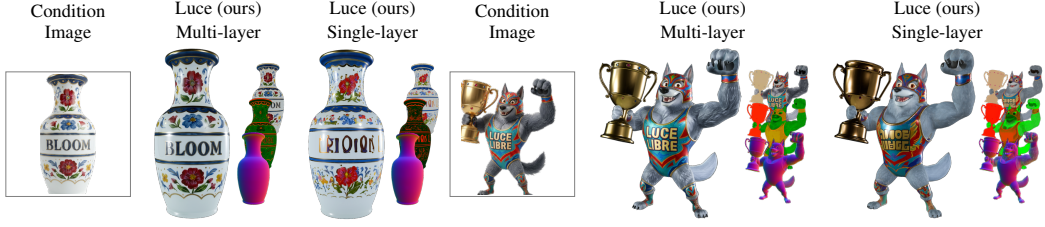

\paragraph{Multi-layer image conditioning.}
Dense prediction tasks have benefited from fusing features across multiple encoder layers~\citep{long2015fcn,lin2017fpn,zhao2017pspnet,chen2017deeplab,ranftl2021dpt,cheng2022mask2former}. Recent work shows the same effect for frozen ViTs: multi-layer DINOv2 features outperform single-layer ones because early layers carry fine spatial patterns while deep layers carry semantics~\citep{karypidis2024dinoforesight}.
We apply this to 3D generation: we concatenate features from several DINOv2 layers and project them into the conditioning space with a two-layer MLP (Appendix~\ref{app:hyperparams}).
Early-layer features carry fine spatial structure (text, logos, inscriptions); fusing them with deep-layer semantics propagates this detail through SLatFlow to the rendered 3D surface (Fig.~\ref{fig:multi_layer_dino_ablation}).
Multi-layer conditioning outperforms single-layer on both Toys4K FID (20.99 vs.\ 25.21) and CLIP on our AI-generated-image benchmark (0.8519 vs.\ 0.8081; Appendix~\ref{app:dino}).

\paragraph{Dual decoding.}
The primary generation output of \method{} is a multimodal PBR Gaussian cloud produced by the Gaussian decoder, a complete material description that splats directly under any environment map via deferred PBR shading (Eq.~\ref{eq:pbr_shading}), with no mesh extraction or UV unwrapping required.
To enable fair comparison with baselines whose primary output is a mesh~\citep{xiang2024trellis,xiang2025trellis2,chen2024primx}, we decode the same latent into a textured mesh using a FlexiCubes~\citep{shen2023flexicubes} mesh decoder, following TRELLIS~\citep{xiang2024trellis}.
We keep its architecture unchanged and retrain it on our learned latent space. Baking decoded normal Gaussians as tangent-space normal maps onto the extracted mesh (Sec.~\ref{sec:normal_transfer}) improves mesh normal PSNR from 29.5 to 33.0~dB (Sec.~\ref{sec:recon}), recovering high-frequency surface detail at zero polygon-count cost.

\subsection{Tangent-Space Normal Map Transfer}
\label{sec:normal_transfer}

Mesh extraction from a voxel grid bounds geometric frequency by the grid resolution, leaving sub-voxel surface features (engravings, fabric weaves, embossed text) unrepresented in the extracted geometry alone.
A standard remedy in real-time rendering is to include a normal map alongside the mesh: this perturbs the geometric normal for lighting calculations, restoring apparent surface detail without raising polygon count.
Our representation lends itself naturally to this.
Because our PBR Gaussians are decoupled from the mesh, we supervise them against the \emph{effective} normals used for rendering (\ie{}, the perturbed geometric normals after applying the authored normal map), so they learn surface orientation at a frequency finer than the mesh can represent.
At inference, we express the decoded normal Gaussians in the mesh's local tangent frame to obtain a tangent-space normal map on the UV-parameterized surface, preserving high-frequency details at a resolution far beyond the mesh geometry.
The final mesh carries four texture maps: diffuse (albedo), metallic, roughness, and tangent-space normal, giving a complete PBR material description directly importable into production renderers.
We detail the baking pipeline in Appendix~\ref{app:hyperparams}.

\section{Experiments}
\label{sec:experiments}
\label{sec:setup}

\begin{table}[t]
\centering
\caption{\textbf{Image-to-3D generation.}
We evaluate on two benchmarks: Toys4K ($N\!=\!412$, left) and our 130 AI-generated images ($N\!=\!130$, right).
Toys4K reports FID/KID with Inception and DINO~\citep{oquab2024dinov2} backbones (KID and KID$_{\text{dino}}$ $\times 100$); both benchmarks report CLIP~\citep{radford2021clip}, SigLIP2~\citep{tschannen2025siglip2}, ULIP~\citep{xue2024ulip2}, and Uni3D-L~\citep{zhou2023uni3d}.
Time(s) is the mean warm wall-clock time per generation on one H100 (Appendix~\ref{app:inference_steps}).
\method{}~GS renders the decoded PBR Gaussians directly via deferred shading; \method{} mesh rows render a textured mesh extracted from the same latent, with and without tangent-space normal map transfer.
Best in \textbf{bold}, second-best \underline{underlined}; shaded rows are ours.
``---'' marks metrics that do not apply to that render path; ULIP and Uni3D-L are mesh-only.}
\label{tab:generation}
\resizebox{\textwidth}{!}{
\begin{tabular}{l c@{\hskip 4pt}c cccccccc cccc}
\toprule
\multirow{2}{*}{Method}
 & \multirow{2}{*}{\shortstack{Params\\(B)}} & \multirow{2}{*}{\shortstack{Time\\(s)}}
 & \multicolumn{8}{c}{Toys4K ($N\!=\!412$)}
 & \multicolumn{4}{c}{AI-generated images ($N\!=\!130$)} \\
\cmidrule(lr){4-11} \cmidrule(lr){12-15}
 & &
 & FID $\downarrow$ & KID $\downarrow$ & FID$_{\text{dino}}$ $\downarrow$ & KID$_{\text{dino}}$ $\downarrow$ & CLIP $\uparrow$ & SigLIP2 $\uparrow$ & ULIP $\uparrow$ & Uni3D-L $\uparrow$
 & CLIP $\uparrow$ & SigLIP2 $\uparrow$ & ULIP $\uparrow$ & Uni3D-L $\uparrow$ \\
\cmidrule[\lightrulewidth](r){1-3} \cmidrule[\lightrulewidth](lr){4-11} \cmidrule[\lightrulewidth](l){12-15}
TRELLIS GS~\citep{xiang2024trellis}     & 1.70 & 4.78   & 30.75 & 0.212 & 0.109 & \underline{0.0013} & 0.8898 & 0.9164 & --- & --- & 0.8299 & 0.8339 & --- & --- \\
TRELLIS mesh~\citep{xiang2024trellis}   & 1.80 & 19.67  & 32.39 & 0.263 & 0.145 & 0.0039 & 0.8829 & 0.9092 & 0.1672 & 0.3747 & 0.8000 & 0.7958 & 0.1247 & 0.3278 \\
3DTopia-XL~\citep{chen2024primx}        & 1.02 & 39.61  & 83.23 & 2.726 & 0.542 & 0.0306 & 0.7644 & 0.7810 & 0.1418 & 0.2519 & 0.6481 & 0.6458 & 0.1108 & 0.2197 \\
TRELLIS~2~\citep{xiang2025trellis2}     & 7.48 & 199.55 & 29.22 & 0.165 & 0.129 & 0.0022 & 0.8895 & 0.9161 & 0.1634 & 0.3635 & 0.8110 & 0.8166 & 0.1207 & 0.3280 \\
LiTo~\citep{chang2026lito}              & 1.84 & 3.75   & 29.76 & 0.208 & 0.128 & 0.0025 & 0.8909 & 0.9082 & --- & --- & 0.8234 & 0.8240 & --- & --- \\
\cmidrule[\lightrulewidth](r){1-3} \cmidrule[\lightrulewidth](lr){4-11} \cmidrule[\lightrulewidth](l){12-15}
\oursrow \method{} GS                               & 4.48 & 7.23   & \textbf{20.99} & \textbf{0.033} & \underline{0.100} & 0.0028 & \underline{0.9062} & 0.9230 & --- & --- & \textbf{0.8519} & \textbf{0.8508} & --- & --- \\
\oursrow \method{} mesh without tangent normal      & 4.58 & 67.14  & 25.04 & 0.102 & 0.112 & 0.0019 & 0.8990 & \underline{0.9234} & \textbf{0.1681} & \underline{0.3767} & 0.8505 & \underline{0.8466} & \textbf{0.1254} & \textbf{0.3298} \\
\oursrow \method{} mesh with tangent normal         & 4.58 & 75.69  & \underline{21.10} & \underline{0.042} & \textbf{0.088} & \textbf{0.0006} & \textbf{0.9072} & \textbf{0.9288} & \underline{0.1679} & \textbf{0.3771} & \underline{0.8508} & 0.8465 & \underline{0.1251} & \underline{0.3295} \\
\bottomrule
\end{tabular}
}
\end{table}

\paragraph{Datasets and evaluation benchmarks.}
Following prior generative 3D works~\citep{xiang2025trellis2} for fair comparison, our training set combines ${\sim}$500K PBR-filtered assets from Objaverse~\citep{deitke2023objaverse} and Objaverse-XL~\citep{deitke2024objaversexl} with a 158K-asset PBR subset of TexVerse~\citep{zhang2025texverse}.
To evaluate, we use \textbf{Toys4K}~\citep{stojanov2021toys4k} for reconstruction, which provides ground-truth 3D assets (a 338-asset PBR subset with all three PBR textures); for generation, we test on \textbf{412 Toys4K assets} and, to gauge generalization beyond Toys4K's simple toy domain, \textbf{130 AI-generated images} from Gemini~3.1 Flash Image~\citep{googledeepmind2026gemini31flashimage}, prompted to be diverse and detail-rich (text, logos, mixed materials).

\paragraph{Baselines.}
We compare against TRELLIS~\citep{xiang2024trellis}, TRELLIS~2~\citep{xiang2025trellis2}, LiTo~\citep{chang2026lito}, and 3DTopia-XL~\citep{chen2024primx} (parameters in Table~\ref{tab:generation}).
All baselines use their public-release image conditioning model and input resolution. TRELLIS and LiTo use DINOv2 ViT-L/14~\citep{oquab2024dinov2}, and 3DTopia-XL DINOv2 ViT-B/14, all at $518\times518$; TRELLIS~2 uses DINOv3~\citep{simeoni2025dinov3} ViT-L/16 at $512\times512$ for its sparse-structure stage and $1024\times1024$ for its latent stage. \method{} (ours) uses DINOv2 ViT-L/14 at $1036\times1036$ for its SLatFlow (Table~\ref{tab:hyperparams}); its sparse-structure stage is inherited unchanged from TRELLIS~2 (Sec.~\ref{sec:flow}).
Baselines with explicit PBR outputs (TRELLIS~2, 3DTopia-XL) are evaluated on shaded color and all three modalities; those without (TRELLIS, LiTo), on shaded color and, where available, surface normals.

\paragraph{Metrics.}
\emph{Generation:} On Toys4K, we report FID~\citep{heusel2017gans} and KID~\citep{binkowski2018demystifying} on rendered views with both Inception and DINO~\citep{oquab2024dinov2} backbones (FID$_{\text{dino}}$, KID$_{\text{dino}}$) as distribution-level metrics. Both benchmarks report alignment metrics: CLIP score, SigLIP2~\citep{tschannen2025siglip2}, ULIP~\citep{xue2024ulip2}, and Uni3D-L, the Large variant of Uni3D~\citep{zhou2023uni3d}, measuring input-output agreement.
\emph{Reconstruction:} We provide per-modality PSNR, SSIM~\citep{wang2004image}, and LPIPS~\citep{zhang2018perceptual} for albedo, metallic-roughness, normal maps, and shaded renders, measuring surface orientation and material reconstruction.
We detail the evaluation rendering and per-benchmark illumination in Appendix~\ref{app:eval_render}.

\subsection{Implementation Details}
\label{sec:implementation}
Model architecture, training hyperparameters, and the texture-baking pipeline are in Appendix~\ref{app:hyperparams}, and inference details in Appendix~\ref{app:inference_steps}; we summarize preprocessing here.

\paragraph{Preprocessing.}
Each training asset is centered and scale-normalized to a unit bounding box, yielding a consistent canonical frame.
We then render each from 150 cameras uniformly distributed on a sphere, producing per-view albedo, metallic-roughness, and normal images at a resolution of $1024\times1024$ pixels. The normal pass is rendered with the asset's authored normal maps applied, so its normals carry surface detail finer than the base geometry.
Per-modality 3DGS is fit independently on each set and voxelized onto a $128^3$ sparse grid with $K\!=\!8$ Gaussians per voxel per modality.
For each modality, we optimize all Gaussians via differentiable rendering against the intrinsic images, parameterizing each center as a $\tanh$-bounded offset from its parent voxel center, initialized at evenly spread quasi-random positions inside the voxel.

\subsection{Generation}
\label{sec:generation}

\begin{figure}[t]
\begin{center}
\input{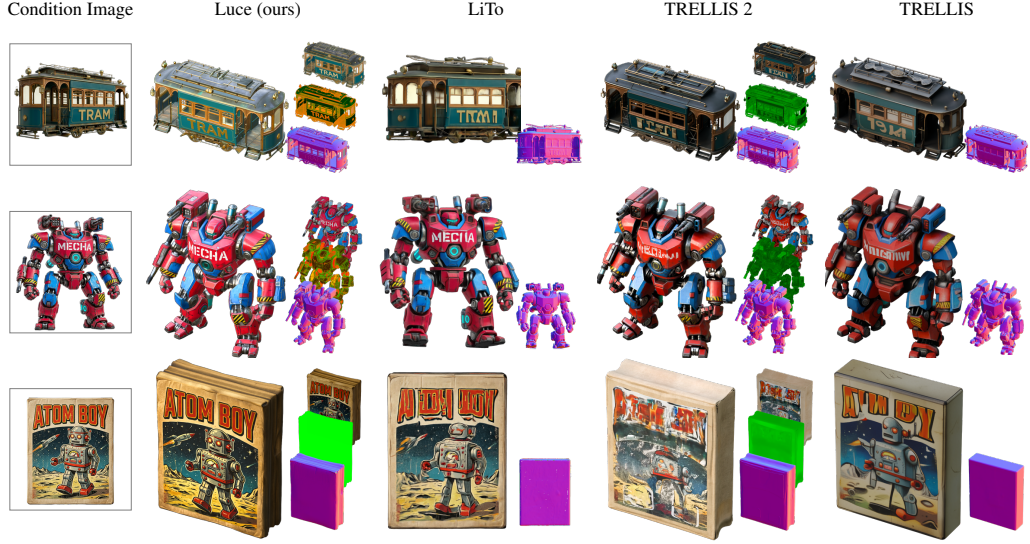}
\end{center}
\vspace{-0.3cm}
\caption{\textbf{Image-to-3D generation comparison.}
\method{} preserves legible text and fine surface details on generated 3D assets. For each method, the larger shaded render is shown next to a cascade of the per-modality decomposition (albedo, metallic-roughness, surface normals) when available.}
\label{fig:gen_compare}
\end{figure}

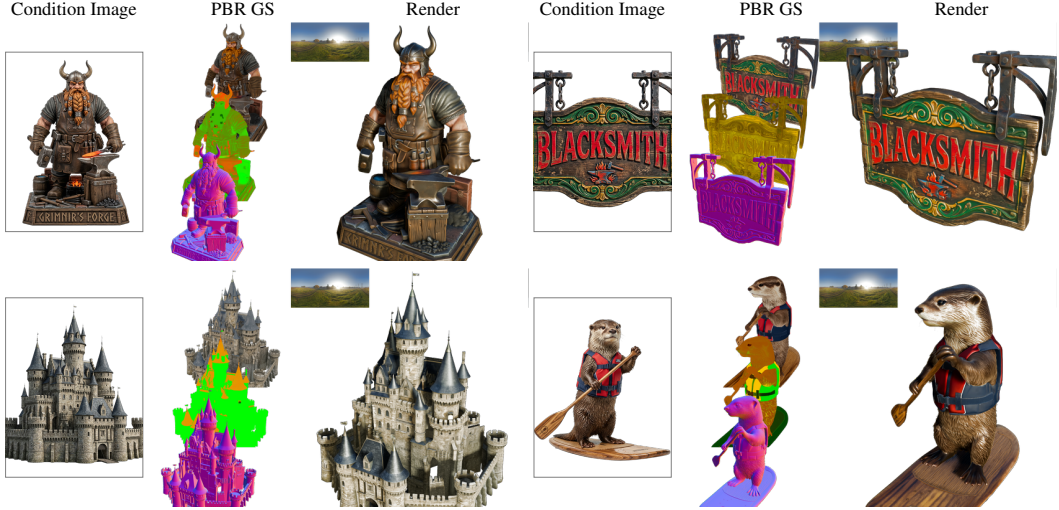
\begin{figure}[t]
\begin{center}
\newcommand{\genqualcell}[1]{%
\begin{tikzpicture}[inner sep=0pt, outer sep=0pt]
  \node[anchor=north west] (strip) at (0,0)
       {\includegraphics[width=0.360\linewidth,trim=1024 0 2048 0,clip]{#1}};
  \node[anchor=east, draw=gray, line width=0.25pt, xshift=-2pt] at (strip.west)
       {\includegraphics[width=0.130\linewidth,trim=120 0 3806 0,clip]{#1}};
\end{tikzpicture}%
}
\newcommand{\genqualheader}{%
\makebox[0.130\linewidth][c]{\scriptsize Condition Image}\hspace{2pt}%
\makebox[0.180\linewidth][c]{\scriptsize PBR GS}%
\makebox[0.180\linewidth][c]{\scriptsize Render}%
}
\setlength{\tabcolsep}{1pt}
\renewcommand{\arraystretch}{0.9}
\begin{tabular}{@{}cc@{}}
\genqualheader & \genqualheader \\
\genqualcell{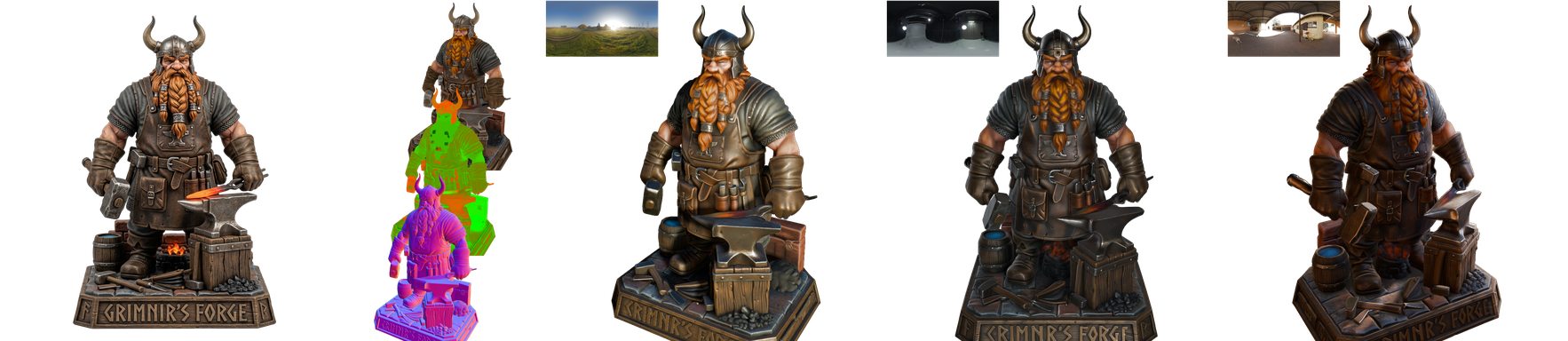} &
\genqualcell{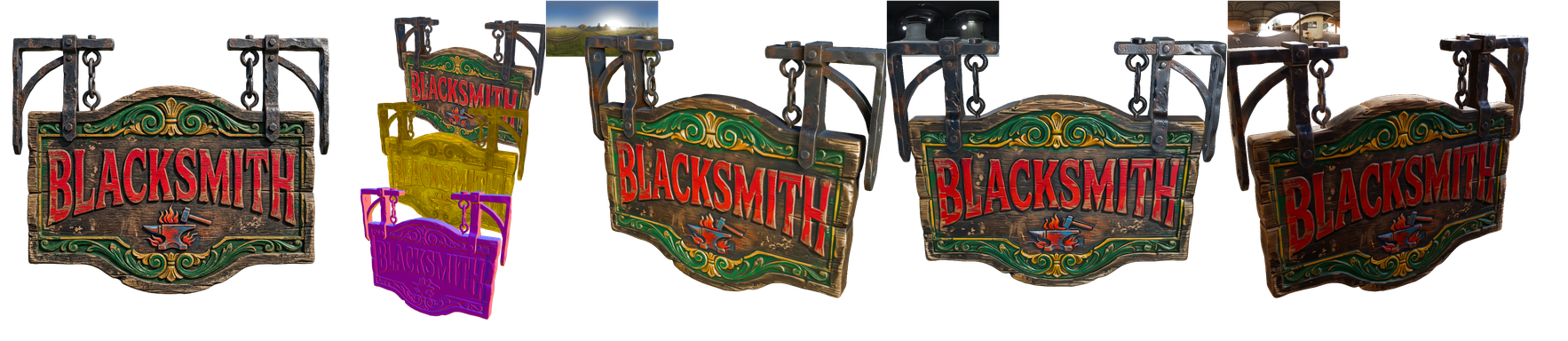} \\
\genqualcell{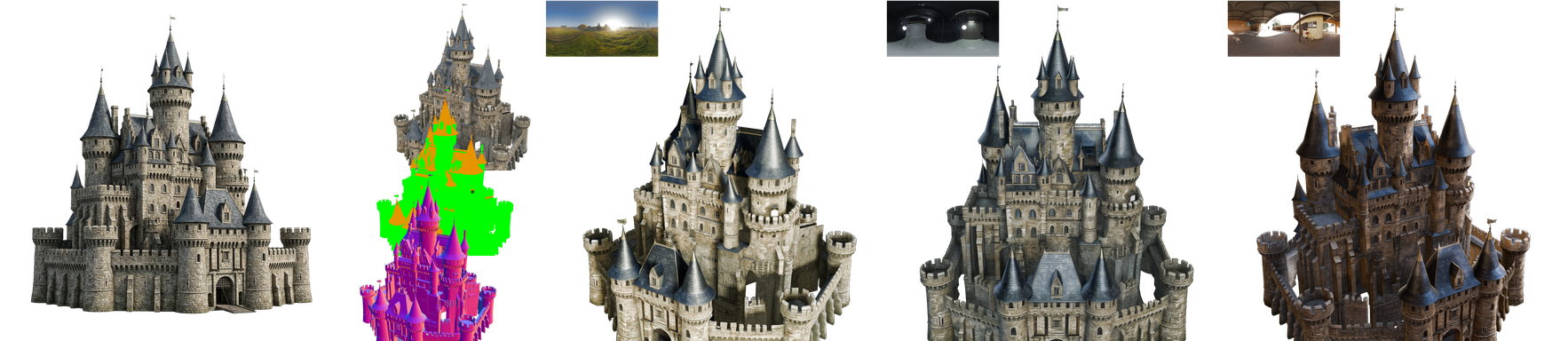} &
\genqualcell{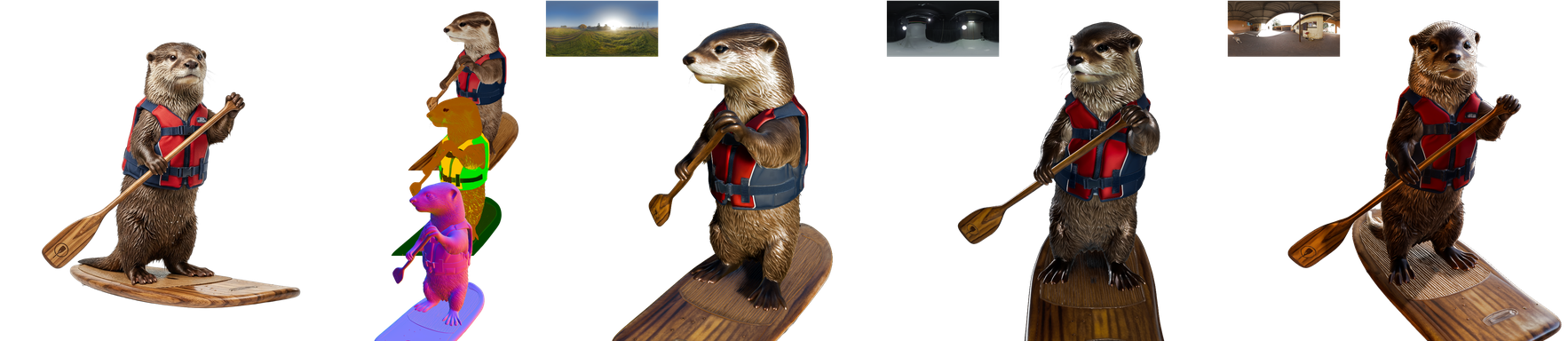} \\
\end{tabular}
\end{center}
\vspace{-0.3cm}
\caption{\textbf{Additional generation examples.}
\method{} on four diverse inputs: each cell shows, from left to right, the condition image, the per-modality PBR Gaussians (albedo, metallic-roughness, normals), and a shaded render under the environment map previewed above it.}
\label{fig:gen_qual}
\end{figure}

Table~\ref{tab:generation} reports generation quality on Toys4K (left) and our AI-generated images (right), with qualitative comparisons against baselines in Fig.~\ref{fig:gen_compare} and additional \method{} examples in Fig.~\ref{fig:gen_qual} (per-sample baseline mosaics in Appendix~\ref{app:baseline_compare}, more examples in Appendix~\ref{app:additional_qual}).
On Toys4K, \method{}~GS reaches the lowest FID (20.99), improving on the strongest baseline TRELLIS~2 (29.22) by more than 8 FID. On our AI-generated-image benchmark, \method{}~GS leads CLIP (0.8519) and SigLIP2 (0.8508) over the best baseline TRELLIS~GS (0.8299 and 0.8339), and the mesh variants lead the mesh-only ULIP and Uni3D-L.
Beyond these alignment wins, \method{} produces per-modality material decomposition natively and supports dual decoding to both Gaussians and textured meshes. The generated assets are relightable under novel illumination (Fig.~\ref{fig:teaser}).
\method{} reproduces legible text and fine surface detail (Fig.~\ref{fig:gen_compare}) that baselines often blur or distort.

\subsection{Reconstruction}
\label{sec:recon}

\begin{table}[t]
\centering
\caption{\textbf{Reconstruction quality} on a PBR subset of Toys4K ($N\!=\!338$).
We report per-modality PSNR, SSIM, and LPIPS for color rendering (combined appearance under fixed illumination), albedo (diffuse color), metallic-roughness, and normal maps (surface orientation).
For \method{}, the \textbf{GS} row renders the decoded PBR Gaussians directly via deferred shading (no mesh); the \textbf{mesh} rows render a textured mesh from the same latent, with and without baked tangent-space normals.
Best in \textbf{bold}, second-best \underline{underlined}; shaded rows are ours.
``---'' indicates that the method does not produce that modality. Params is the autoencoder parameter count in billions.}
\label{tab:recon_appearance}
\setlength{\tabcolsep}{8pt}%
\resizebox{\textwidth}{!}{
\begin{tabular}{l ccc ccc ccc ccc c}
\toprule
\multirow{2}{*}{Method} & \multicolumn{3}{c}{Color} & \multicolumn{3}{c}{Albedo} & \multicolumn{3}{c}{Metallic-Roughness} & \multicolumn{3}{c}{Normal} & \multirow{2}{*}{\shortstack{Params\\(B)}} \\
\cmidrule(lr){2-4} \cmidrule(lr){5-7} \cmidrule(lr){8-10} \cmidrule(lr){11-13}
& PSNR$\uparrow$ & SSIM$\uparrow$ & LPIPS$\downarrow$ & PSNR$\uparrow$ & SSIM$\uparrow$ & LPIPS$\downarrow$ & PSNR$\uparrow$ & SSIM$\uparrow$ & LPIPS$\downarrow$ & PSNR$\uparrow$ & SSIM$\uparrow$ & LPIPS$\downarrow$ & \\
\midrule
TRELLIS GS~\citep{xiang2024trellis}     & 27.0 & 0.912 & 0.124 & ---  & ---   & ---   & ---  & ---   & ---   & ---  & ---   & ---   & 0.17 \\
TRELLIS mesh~\citep{xiang2024trellis}   & 26.3 & 0.907 & 0.118 & ---  & ---   & ---   & ---  & ---   & ---   & 26.9 & 0.914 & 0.113 & 0.26 \\
3DTopia-XL~\citep{chen2024primx}        & 23.9 & 0.877 & 0.188 & 25.4 & 0.893 & 0.191 & 22.6 & 0.910 & 0.191 & 23.3 & 0.876 & 0.165 & 0.03 \\
TRELLIS~2~\citep{xiang2025trellis2}     & \underline{34.5} & \underline{0.961} & \underline{0.074} & \textbf{40.7} & \textbf{0.976} & \textbf{0.051} & \textbf{42.7} & \textbf{0.986} & \textbf{0.038} & 32.1 & 0.940 & 0.093 & 1.66 \\
LiTo~\citep{chang2026lito}              & 31.9 & 0.944 & 0.110 & ---  & ---   & ---   & ---  & ---   & ---   & 27.6 & 0.912 & 0.116 & 0.29 \\
\midrule
\oursrow \method{} GS                                  & \textbf{36.1} & \textbf{0.967} & \textbf{0.059} & \underline{38.6} & \underline{0.971} & \underline{0.055} & \underline{39.1} & \underline{0.978} & \underline{0.048} & \textbf{34.6} & \textbf{0.961} & \textbf{0.038} & 1.82 \\
\oursrow \method{} mesh without tangent normal         & 31.1 & 0.944 & 0.091 & 33.7 & 0.961 & 0.067 & 37.9 & 0.977 & 0.054 & 29.5 & 0.928 & 0.103 & 1.91 \\
\oursrow \method{} mesh with tangent normal            & 32.5 & 0.953 & 0.080 & 33.7 & 0.961 & 0.067 & 37.9 & 0.977 & 0.054 & \underline{33.0} & \underline{0.952} & \underline{0.067} & 1.91 \\
\bottomrule
\end{tabular}
}
\end{table}

\begin{figure}[t]
\centering
\gdef\sp{ }
\def\reconzoomXa{185}\def\reconzoomYa{640}\def\reconzoomXb{485}\def\reconzoomYb{750}
\newcommand{\reconcell}[2]{%
\begin{minipage}[t]{0.24\linewidth}\centering
{\scriptsize\strut #1}\\[2pt]%
\begin{tikzpicture}[inner sep=0pt, outer sep=0pt]
  \node[anchor=north west] (cell) at (0,0)
       {\includegraphics[width=\linewidth]{#2}};
  \draw[green, line width=0.4pt]
    ($(cell.north west)+({(\reconzoomXa/1369)*\linewidth},{-(\reconzoomYa/1369)*\linewidth})$) rectangle
    ($(cell.north west)+({(\reconzoomXb/1369)*\linewidth},{-(\reconzoomYb/1369)*\linewidth})$);
  \edef\Lz{\reconzoomXa}%
  \edef\Rz{\the\numexpr 1369-\reconzoomXb\relax}%
  \edef\Tz{\reconzoomYa}%
  \edef\Bz{\the\numexpr 1024-\reconzoomYb\relax}%
  \edef\figIGcall{\noexpand\includegraphics[width=\linewidth, trim={\Lz\sp\Bz\sp\Rz\sp\Tz}, clip]{#2}}%
  \node[anchor=north west, draw=green, line width=0.4pt, inner sep=0pt]
       at ($(cell.south west)+(0,-2pt)$) {\figIGcall};
\end{tikzpicture}
\end{minipage}%
}
\reconcell{GT}{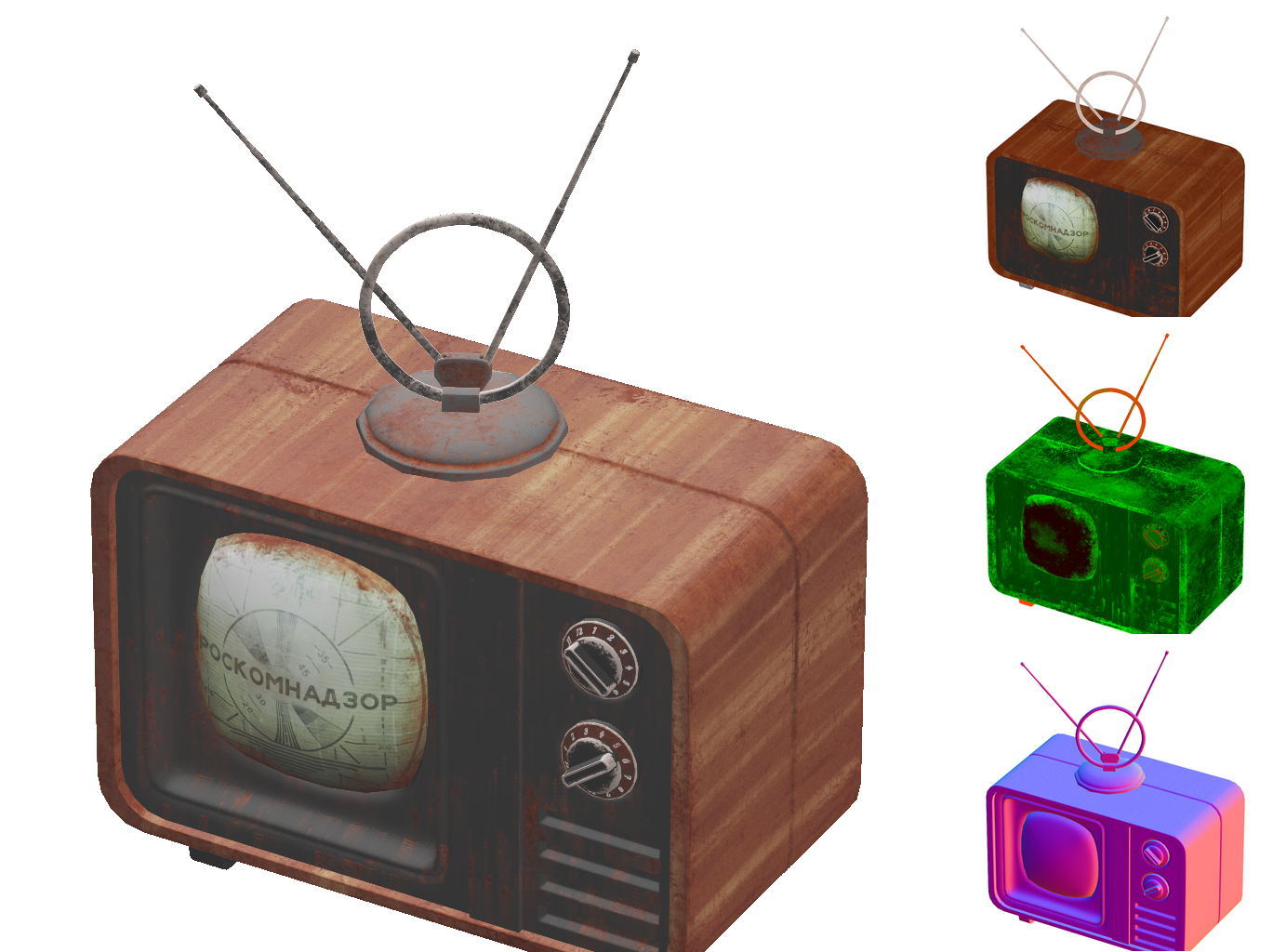}\hfill
\reconcell{\method{}~GS (ours)}{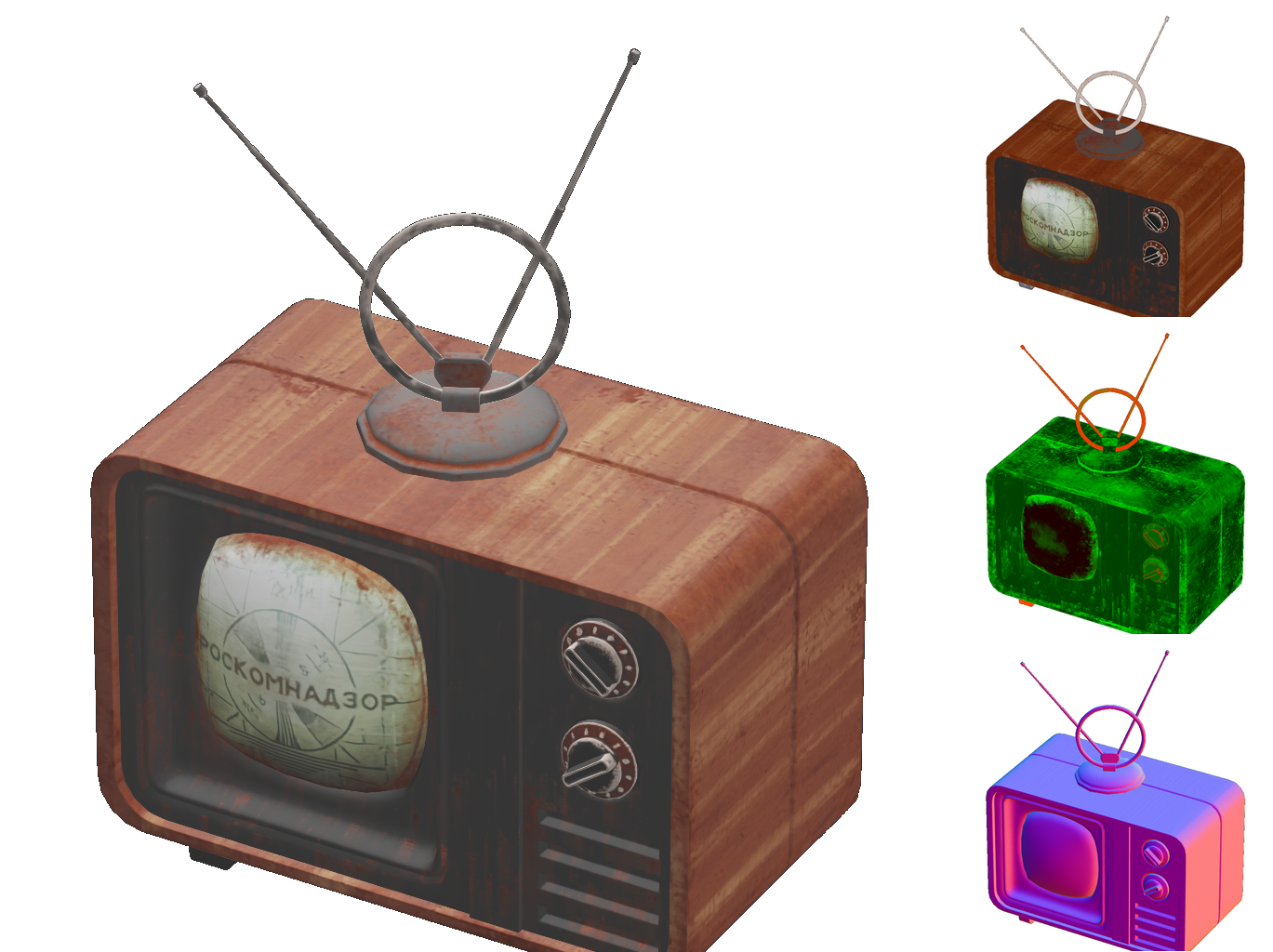}\hfill
\reconcell{TRELLIS~2}{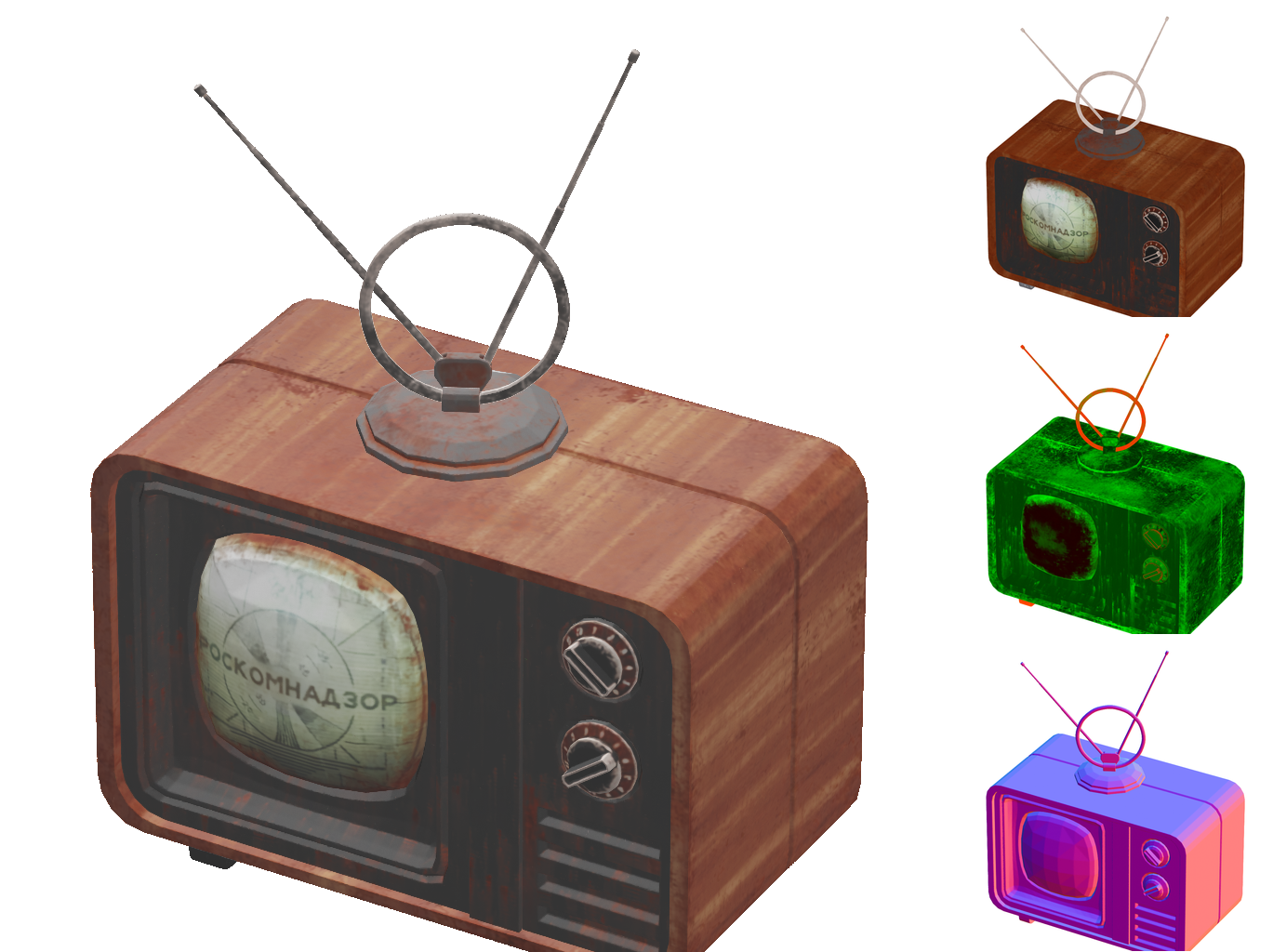}\hfill
\reconcell{TRELLIS}{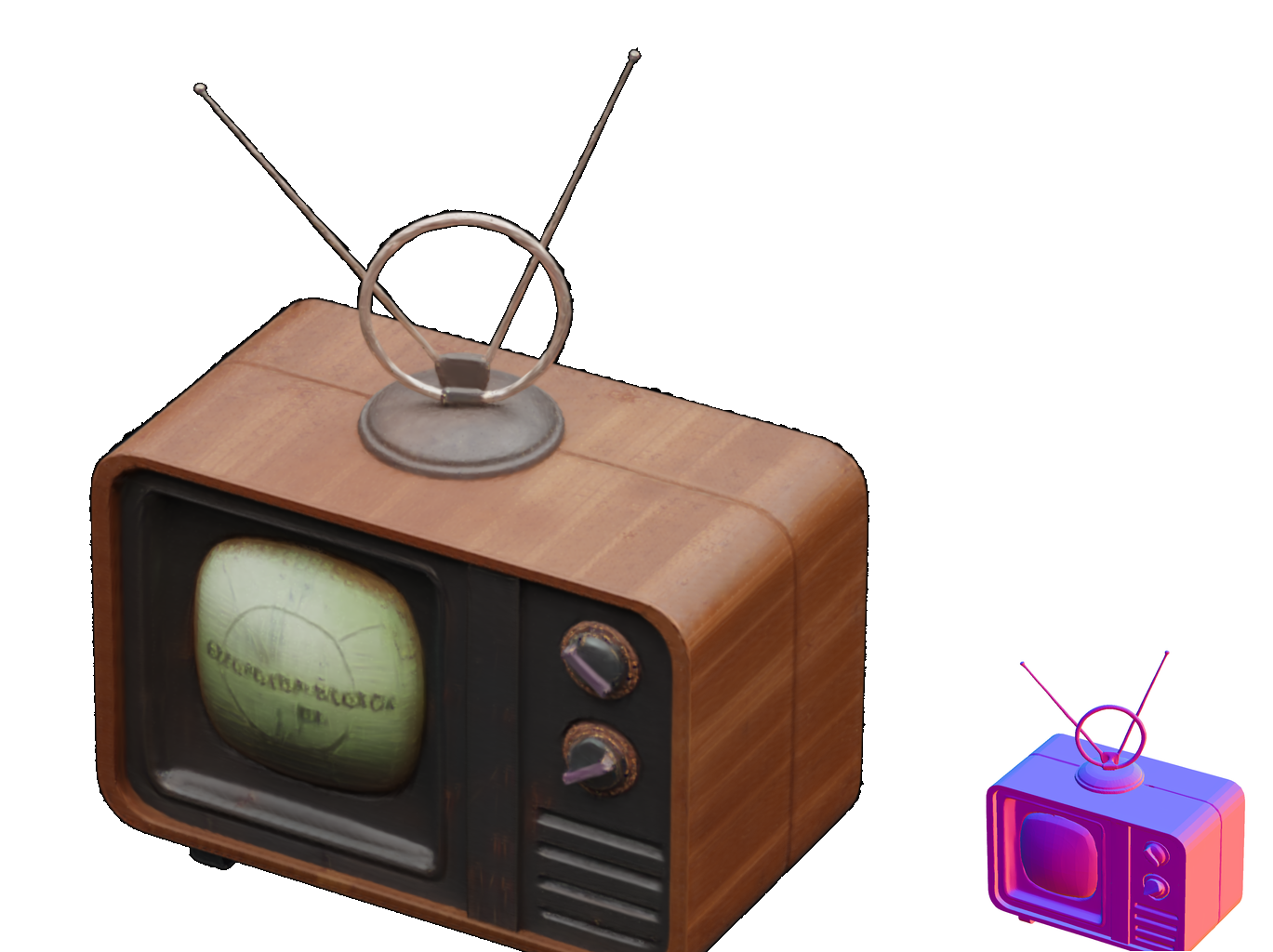}
\vspace{-0.3cm}
\caption{\textbf{Reconstruction quality on Toys4K.}
Columns are ground truth and methods; each block shows the modalities that each method reconstructs (albedo, metallic-roughness, normals).
\method{} preserves fine detail (text, highlights, texture); the green box marks the region magnified below.}
\label{fig:recon_qual}
\end{figure}
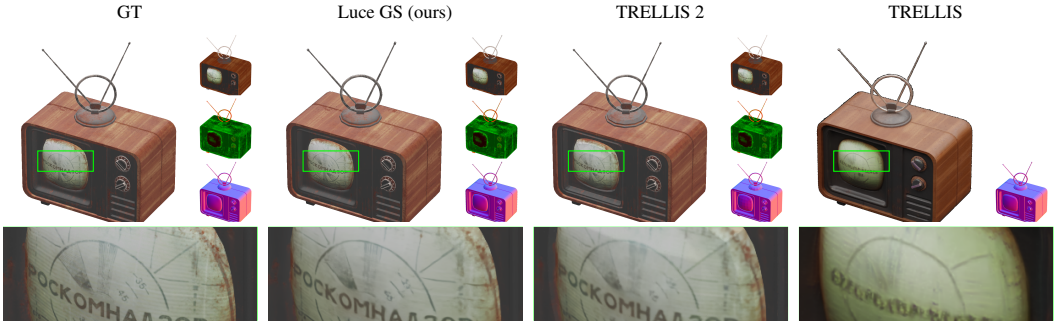

\gdef\sp{ }
\gdef\nmlzoom#1#2#3#4#5#6{%
  \edef\Lz{\the\numexpr(#3)\relax}%
  \edef\Bz{\the\numexpr 1024-(#6)\relax}%
  \edef\Rz{\the\numexpr 1024-(#5)\relax}%
  \edef\Tz{\the\numexpr(#4)\relax}%
  \edef\nmlIG{\noexpand\includegraphics[width=#2, trim={\Lz\sp\Bz\sp\Rz\sp\Tz}, clip]{#1}}%
  \nmlIG
}
\begin{figure}[t]
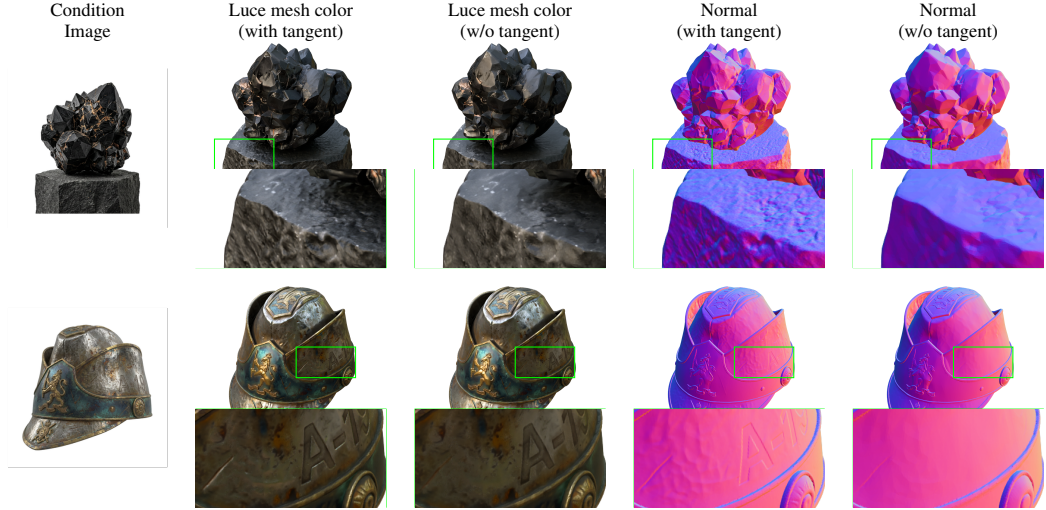

\centering
\begin{tikzpicture}[inner sep=0pt, outer sep=0pt]
\def\cw{0.18\linewidth}
\def\condw{0.15\linewidth}
\def\gp{0.0275\linewidth}
\def\lbldrop{0.026\linewidth}

\foreach \r/\id/\rk/\cxa/\cya/\cxb/\cyb in {%
  1/096/0/100/675/420/840,
  2/128/-1.26/540/505/860/670%
} {
  \node[anchor=north west, draw=gray, line width=0.25pt] at (0, {\rk*\cw-0.707*\cw+0.5*\condw})
    {\includegraphics[width=\condw]{figures/gen_normal_compare/image_\id/cond.png}};
  \foreach \c/\suf in {1/color_tg, 2/color_notg, 3/normal_tg, 4/normal_notg} {
    \node[anchor=north west] (cellN\r\c) at ({\condw+\gp+(\c-1)*(\cw+\gp)}, {\rk*\cw})
      {\includegraphics[width=\cw]{figures/gen_normal_compare/image_\id/\suf.png}};
    \draw[green, line width=0.4pt]
      ($(cellN\r\c.north west)+({(\cxa/1024)*\cw}, {-(\cya/1024)*\cw})$) rectangle
      ($(cellN\r\c.north west)+({(\cxb/1024)*\cw}, {-(\cyb/1024)*\cw})$);
  }
}
\foreach \r/\id/\rk/\cxa/\cya/\cxb/\cyb in {%
  1/096/0/100/675/420/840,
  2/128/-1.26/540/505/860/670%
} {
  \foreach \c/\suf in {1/color_tg, 2/color_notg, 3/normal_tg, 4/normal_notg} {
    \node[anchor=north west, draw=green, line width=0.4pt, inner sep=0pt]
      at ($(cellN\r\c.south west)+(0,{0.18*\cw})$)
      {\nmlzoom{figures/gen_normal_compare/image_\id/\suf.png}{\cw}{\cxa}{\cya}{\cxb}{\cyb}};
  }
}
\node[anchor=south, font=\scriptsize, align=center] at ({0.5*\condw}, -\lbldrop) {Condition\\Image};
\node[anchor=south, font=\scriptsize, align=center] at ({\condw+\gp+0.5*\cw}, -\lbldrop) {\method{} mesh color\\(with tangent)};
\node[anchor=south, font=\scriptsize, align=center] at ({\condw+2*\gp+1.5*\cw}, -\lbldrop) {\method{} mesh color\\(w/o tangent)};
\node[anchor=south, font=\scriptsize, align=center] at ({\condw+3*\gp+2.5*\cw}, -\lbldrop) {Normal\\(with tangent)};
\node[anchor=south, font=\scriptsize, align=center] at ({\condw+4*\gp+3.5*\cw}, -\lbldrop) {Normal\\(w/o tangent)};
\end{tikzpicture}
\caption{\textbf{Tangent-space normal map transfer.}
\method{} bakes tangent-space normals from decoded normal Gaussians onto the extracted mesh, recovering fine surface detail at zero polygon-count cost. Each green box marks the region magnified below.}
\label{fig:normal_transfer}
\end{figure}

We report reconstruction as supporting evidence for our representation choices; generation (Sec.~\ref{sec:generation}) remains our primary contribution, exercising the latent end-to-end.
Table~\ref{tab:recon_appearance} reports per-modality fidelity, with qualitative examples in Fig.~\ref{fig:recon_qual}.
\method{}~GS achieves the best color rendering (36.1~dB PSNR) and normal reconstruction (34.6~dB), surpassing TRELLIS~2, which leads on albedo and metallic-roughness, though \method{}'s albedo LPIPS is close (0.055 vs.\ 0.051).

\subsection{Ablation Studies}
\label{sec:ablations}

We run two ablations. First, we ablate image conditioning by training \method{} with single-layer instead of multi-layer DINOv2 features; multi-layer improves all metrics (Appendix~\ref{app:dino}, Table~\ref{tab:ablation_dino}). Second, for tangent-space normal map transfer, baking the decoded normal Gaussians onto the mesh (Sec.~\ref{sec:normal_transfer}) improves normal fidelity across all metrics and enhances color reconstruction (Table~\ref{tab:recon_appearance}). Figure~\ref{fig:normal_transfer} compares each example with and without normal baking using shaded color and surface-normal renderings. Baking recovers fine surface details, such as rough textures, engravings, and dents, that are smoothed out by the bare mesh.

\section{Conclusion}
\label{sec:conclusion}

We presented \method{}, a multimodal PBR Gaussian representation for relightable image-to-3D generation.
By giving each occupied voxel its own Gaussian sets for albedo, metallic-roughness, and normals, the representation makes materials explicit in a format compatible with both flow-based generation and production rendering.
The learned latent is compact and diffusible, and decodes into relightable PBR Gaussians and textured meshes with tangent-space normal maps.
Across standard benchmarks, these design choices yield state-of-the-art generation quality while producing relightable outputs by design. On Toys4K, \method{} achieves an FID of 20.99, improving on the strongest baseline TRELLIS~2 by more than 8 FID, and leads CLIP and SigLIP2 alignment on our AI-generated-image benchmark. Together these results help bridge generative 3D modeling and production-compatible asset creation.

\paragraph{Limitations and future work.}
\method{} inherits several limitations from its finite-resolution voxelized representation. Assets whose detail is fine relative to their overall extent may be under-resolved, with features spanning only a few voxels. A natural extension is to use cascaded or adaptive-resolution decoding, where a coarse latent captures global shape and material structure while higher-resolution stages refine local geometry, textures, and normals. Our current material model focuses on standard PBR attributes (albedo, metallic-roughness, and normals) under a Cook--Torrance reflectance model. While this covers many common asset types, it does not explicitly model more complex appearance effects such as subsurface scattering, anisotropy, translucency, thin-film interference, or strongly view-dependent reflectance. Extending the representation with additional material channels or learned view-dependent residuals could further improve realism for challenging materials. Our mesh export uses the FlexiCubes~\citep{shen2023flexicubes} decoder architecture from TRELLIS~\citep{xiang2024trellis}; improving it with higher-resolution or learned UV-aware extraction is a direction for future work. Finally, our pipeline targets object-centric assets; extending \method{} to scene-level generation with coherent lighting, material consistency, and object interactions remains an open direction.

\FloatBarrier

\section*{Acknowledgements}
We are grateful to Jen-Hao Rick Chang, Miguel Angel Bautista Martin, Nafees Bin Zafar, and Barry-John Theobald for their valuable discussion and feedback on our paper. We also thank Federico Semeraro and the broader Apple infrastructure team for maintaining the computing resources that supported this work.

\bibliographystyle{preprint}
\bibliography{references}

\clearpage
\appendix

\begin{center}{\large\bf Appendix}\end{center}
\vspace{0.5em}
\noindent Appendix~\ref{app:hyperparams} details the model architecture, training setup, and
texture-baking pipeline. Appendix~\ref{app:dino} isolates the multi-layer conditioning ablation,
and Appendix~\ref{app:inference_steps} reports inference cost and sampling steps.
Appendix~\ref{app:eval_render} specifies the evaluation renders, illumination, and FID/KID protocol.
Appendices~\ref{app:additional_qual} and~\ref{app:baseline_compare} give additional qualitative
results and per-sample comparisons against every baseline.

\section{Model Architecture and Implementation Details}
\label{app:hyperparams}

\begin{table}[h]
\centering
\caption{\textbf{SLatVAE and SLatFlow hyperparameters.} Architecture diagram in Fig.~\ref{fig:vae_arch}.}
\label{tab:hyperparams}
\begin{tabular}{lcc}
\toprule
& SLatVAE & SLatFlow \\
\midrule
\multicolumn{3}{l}{\textit{Architecture}} \\
\quad Transformer blocks & 16 & 30 \\
\quad Model channels & 1536 & 1536 \\
\quad I/O block channels & -- & [1024] \\
\quad Attention heads & 12 & 12 \\
\quad MLP ratio & 4 & 6 \\
\quad Attention mode & Swin (window 8) & Full (flash varlen) \\
\quad Input channels & 336 & 32 \\
\quad Latent / output channels & 32 & 32 \\
\quad Position encoding & -- & RoPE~\citep{su2024roformer} \\
\quad adaLN modulation & -- & 6-way (shift, scale, gate $\times$ 2) \\
\midrule
\multicolumn{3}{l}{\textit{Training}} \\
\quad GPUs & 64 (H100) & 64 (H100) \\
\quad Batch size (per GPU) & 2 & 3 \\
\quad Max voxels per batch & 400K & 8M \\
\quad Training steps & 500K & 500K \\
\quad Wall-clock time & ${\sim}$14 days & ${\sim}$14 days \\
\quad EMA decay & 0.9999 & 0.9999 \\
\quad Precision & BF16 & BF16 \\
\midrule
\multicolumn{3}{l}{\textit{Data \& conditioning}} \\
\quad Image resolution & $1024\times1024$ (render) & $1036\times1036$ (cond.) \\
\quad Background & random & 0.5 \\
\quad Crop padding & -- & 1.2 \\
\midrule
\multicolumn{3}{l}{\textit{Loss}} \\
\quad Primary & $\ell_1$ + LPIPS + SSIM + KL & smooth $\ell_1$ \\
\quad $\lambda_{\ell_1}$ / $\lambda_{\text{LPIPS}}$ / $\lambda_{\text{SSIM}}$ & 1.0 / 0.2 / 0.2 & -- \\
\quad $\lambda_{\text{KL}}$ & $10^{-3}$ & -- \\
\quad $\lambda_{\text{vol}}$ & $10^{4}$ & -- \\
\quad $\lambda_{o}$ & $10^{-3}$ & -- \\
\quad Timestep sampling & -- & logit-normal ($\mu\!=\!1$, $\sigma\!=\!1$) \\
\midrule
\multicolumn{3}{l}{\textit{Optimizer}} \\
\quad Type & Muon~\citep{jordan2024muon} & Muon~\citep{jordan2024muon} \\
\quad Learning rate & $10^{-3}$ & $10^{-3}$ \\
\quad Weight decay & $10^{-3}$ & $10^{-3}$ \\
\bottomrule
\end{tabular}
\end{table}

\begin{figure}[h]
\centering
\includegraphics[width=\linewidth]{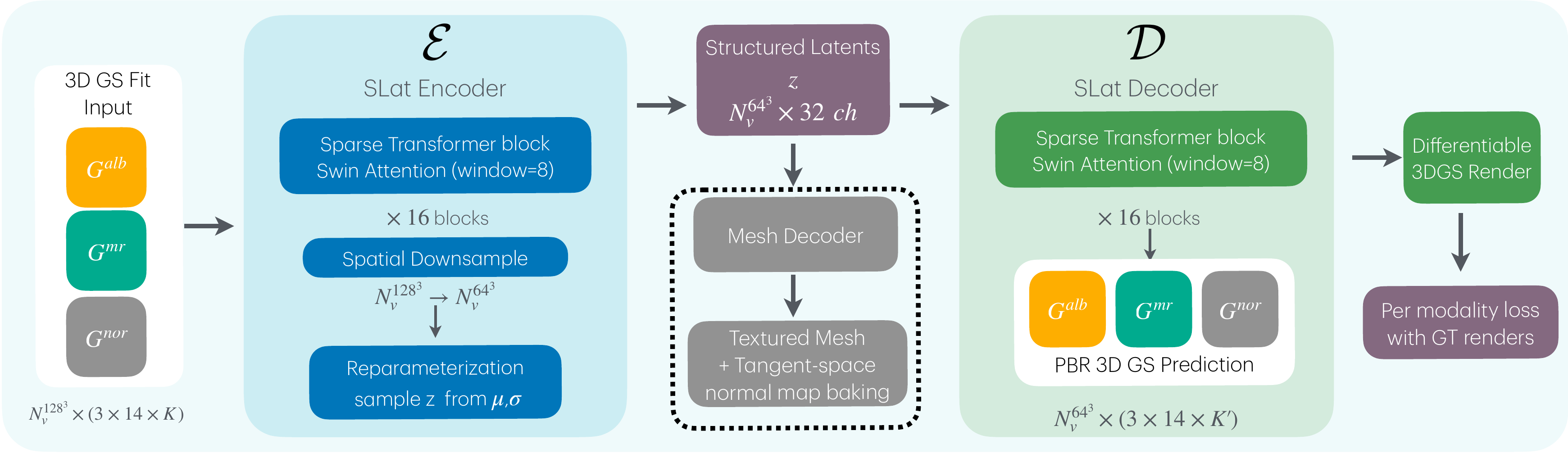}
\caption{\textbf{SLatVAE architecture.}
The encoder compresses the sparse multimodal Gaussian cloud into a compact latent grid.
The decoder reconstructs a denser set of Gaussians per voxel per modality, supervised through per-modality differentiable rendering.
The same latent also serves as input to a mesh decoder for textured mesh extraction.
Full hyperparameters in Table~\ref{tab:hyperparams}.}
\label{fig:vae_arch}
\end{figure}

\begin{figure}[h]
\centering
\setlength{\tabcolsep}{0pt}
\begin{tabular}{@{}cccc@{}}
\makebox[0.25\linewidth][c]{\scriptsize \method{} PBR Gaussians (ours)} &
\makebox[0.25\linewidth][c]{\scriptsize Blender EEVEE} &
\makebox[0.25\linewidth][c]{\scriptsize \method{} PBR Gaussians (ours)} &
\makebox[0.25\linewidth][c]{\scriptsize Blender EEVEE} \\
\multicolumn{4}{@{}c@{}}{\includegraphics[width=\linewidth]{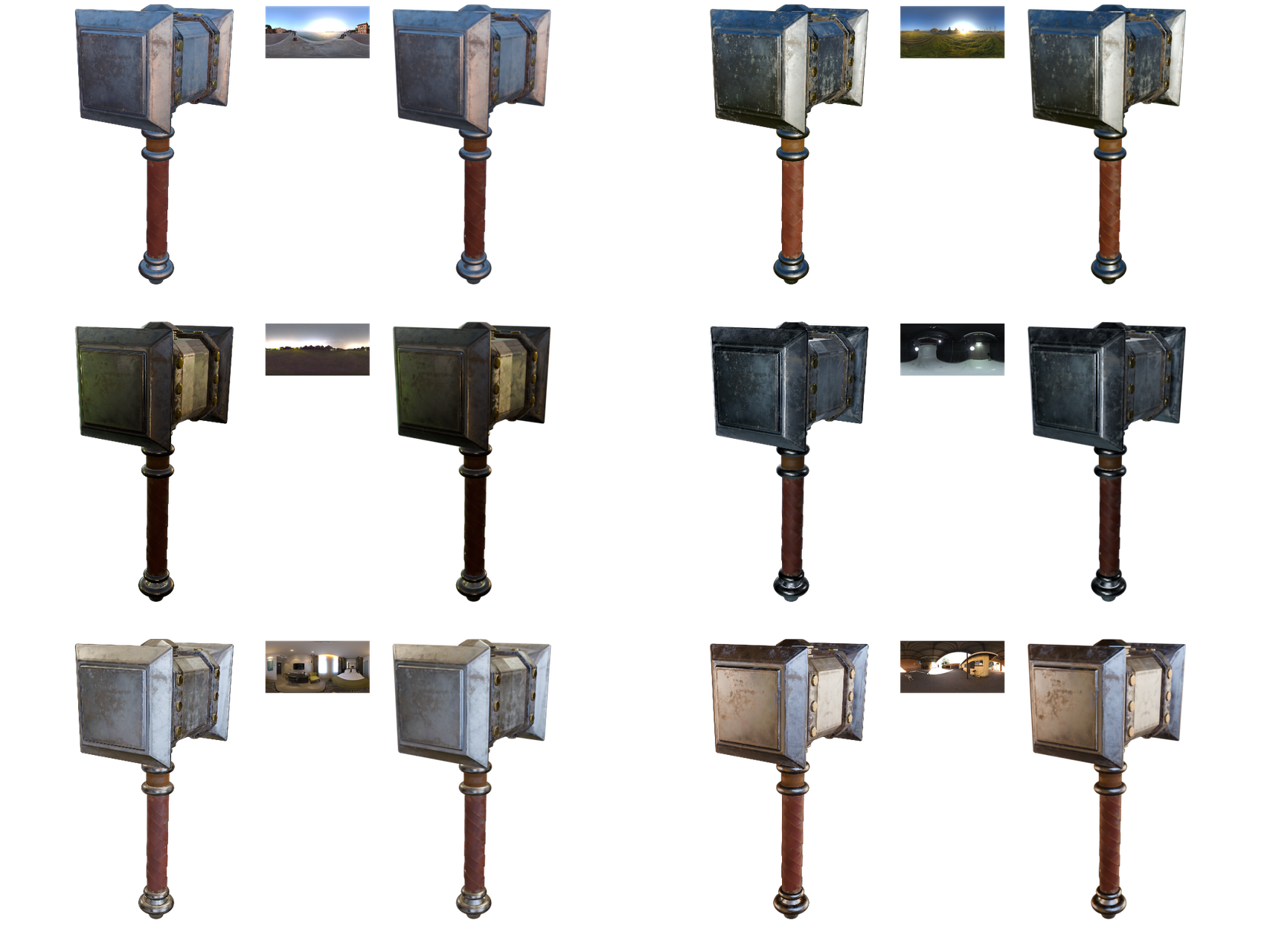}} \\
\end{tabular}
\caption{\textbf{PBR shaded rendering.}
The PBR modalities (albedo, metallic-roughness, normals) enable shaded rendering under novel environment maps~\citep{polyhaven_hdris} via standard PBR compositing (Eq.~\ref{eq:pbr_shading}), without mesh extraction or UV unwrapping. We compare our deferred PBR renderer on the GS fit against the textured mesh rendered with Blender EEVEE~\citep{Blender} as a reference; the environment maps are listed in Appendix~\ref{app:eval_render}.}
\label{fig:vae_relight}
\end{figure}

\paragraph{SLatVAE.}
The encoder is a sparse transformer that operates at the input voxel resolution ($128^3$) throughout its 16 blocks, then performs a single $128^3 \to 64^3$ downsample ($d\!=\!128 \to d'\!=\!64$). This delays compression to the latest possible point in the network, allowing the encoder to reason about fine-grained per-voxel Gaussian features at the input grid before pooling them into the compact latent. The decoder uses the same 16-block sparse transformer but stays at the latent resolution ($64^3$) throughout, emitting $K'\!=\!32$ Gaussians per voxel per modality (versus $K\!=\!8$ at the input) to realize the resolution-for-density trade of Sec.~\ref{sec:vae}.

\paragraph{SLatFlow.}
SLatFlow is a 30-block DiT with rotary position embeddings (RoPE)~\citep{su2024roformer} on integer 3D voxel coordinates and QK-RMSNorm, and applies no internal spatial downsampling or upsampling: the active-voxel token sequence keeps a single fixed resolution across all blocks. To avoid padding, samples within a batch are concatenated into a single flat tensor with per-sample boundary tracking, and self-attention uses variable-length flash attention~\citep{xiang2024trellis} so each sample attends only within its own token set. Each block applies adaptive layer normalization conditioned on the diffusion timestep (six-way modulation). For image conditioning, DINOv2 ViT-L/14 features from layers 6, 12, 18, and 24 are concatenated into 4096 channels ($4 \times 1024$). A LayerNorm followed by a two-layer MLP ($4096 \to 2048$, GELU $\to 1024$) projects them to the 1024-D conditioning space. The result is injected via cross-attention at the fixed token resolution. Layer indices refer to the outputs of the encoder's transformer blocks.

\paragraph{Tensor layout.}
For uniform per-Gaussian dimensionality across modalities, the metallic-roughness modality is stored with a padded 3-channel value (metallic, roughness, and an unused channel), matching the 3-channel albedo (RGB) and normal direction.
The 336-channel SLatVAE input then corresponds to $|\mathcal{M}| \cdot 14 \cdot K = 3 \cdot 14 \cdot 8$, with $K\!=\!8$ Gaussians per modality and 14 parameters per Gaussian (3+3+4+1+3 for position, scale, rotation, opacity, and modality value).

\paragraph{Training objective.}
The SLatVAE is trained end-to-end through differentiable rendering of the decoded Gaussians against ground-truth intrinsic images with randomized, constant-color backgrounds.
Each modality $m \in \mathcal{M} = \{\text{albedo}, \text{normal}, \text{metallic-roughness}\}$ carries its own reconstruction loss, providing separate reconstruction supervision for each modality:
\begin{equation}
    \mathcal{L}_{\text{VAE}} = \sum_{m \in \mathcal{M}} \bigl(\lambda_{\ell_1}\mathcal{L}_{\ell_1}^{m} + \lambda_{\text{LPIPS}}\mathcal{L}_{\text{LPIPS}}^{m} + \lambda_{\text{SSIM}}\mathcal{L}_{\text{SSIM}}^{m}\bigr) + \lambda_{\text{KL}}\,D_{\text{KL}}(q \| p) + \lambda_{\text{vol}}\mathcal{L}_{\text{vol}} + \lambda_{o}\mathcal{L}_{o},
    \label{eq:vae_loss}
\end{equation}
where $\mathcal{L}_{\ell_1}^{m}$, $\mathcal{L}_{\text{LPIPS}}^{m}$, and $\mathcal{L}_{\text{SSIM}}^{m}$ are pixel, perceptual, and structural similarity losses on the rendered images of modality $m$, and $D_{\text{KL}}(q \| p)$ is the KL divergence between the approximate posterior and the latent prior.
The two regularizers operate directly on the predicted Gaussian parameters, with $\mathcal{G}$ the set of all decoded Gaussians in a sample: $\mathcal{L}_{\text{vol}} = \tfrac{1}{|\mathcal{G}|}\sum_{g \in \mathcal{G}} \prod_{i=1}^{3} s^{(g)}_i$ penalizes the volume of each Gaussian (with $\bs^{(g)}$ the per-axis scale) to discourage oversized primitives, and $\mathcal{L}_{o} = \tfrac{1}{|\mathcal{G}|}\sum_{g \in \mathcal{G}} (1 - o^{(g)})$ pulls per-Gaussian opacities $o^{(g)}$ toward one to prevent the decoder from collapsing capacity into transparent primitives.
Loss coefficients are listed in Table~\ref{tab:hyperparams}.

\paragraph{Optimizer.}
We optimize both the SLatVAE and SLatFlow with Muon~\citep{jordan2024muon}, an orthogonalized momentum optimizer for hidden-layer matrix parameters, with AdamW~\citep{loshchilov2017decoupled} used for embeddings, biases, and normalization parameters. Both models use a learning rate and weight decay of $10^{-3}$.

\paragraph{Texture baking.}
After FlexiCubes mesh extraction, we use xatlas to compute a UV parameterization of the surface. For each PBR modality (albedo, metallic-roughness, and normals), we render the decoded Gaussian field from 150 viewpoints and fit a UV-space texture to the rendered targets under a total-variation regularizer. Duplicate vertices are merged so shading normals interpolate smoothly across UV seams. Normals require one additional step: for each occupied UV texel, the baked object-space normal is transformed into tangent space using the mesh tangent frame evaluated at that texel, obtained by rasterizing the per-vertex tangent frames over the UV chart. This produces a standard tangent-space normal map for conventional PBR pipelines (Sec.~\ref{sec:normal_transfer}).

\section{Multi-Layer DINOv2 Conditioning Ablation}
\label{app:dino}

This appendix isolates the contribution of multi-layer DINOv2 conditioning to image-conditioned generation.
We train a single-layer SLatFlow variant that conditions only on layer-24 DINOv2 features and compare against our default that fuses features from layers 6, 12, 18, and 24 (Sec.~\ref{sec:flow}).
All other architecture and training choices are held fixed; both variants are evaluated on the GS render path. The single-layer variant carries no fusion projection.

\begin{table}[h]
\centering
\caption{\textbf{Multi-layer DINOv2 conditioning ablation.}
Single-layer (layer 24) versus multi-layer (layers 6, 12, 18, 24) image conditioning for SLatFlow; both rows use the GS render path, so ULIP and Uni3D-L (mesh-only metrics in our setup) are omitted. KID is reported $\times 100$; FID$_{\text{dino}}$ and KID$_{\text{dino}}$ use a DINOv2 backbone.
Both rows are \method{}; the shaded row is our default.}
\label{tab:ablation_dino}
\resizebox{\textwidth}{!}{
\begin{tabular}{l cccccc cc}
\toprule
\multirow{2}{*}{DINOv2 conditioning}
 & \multicolumn{6}{c}{Toys4K ($N\!=\!412$)}
 & \multicolumn{2}{c}{AI-generated images ($N\!=\!130$)} \\
\cmidrule(lr){2-7} \cmidrule(lr){8-9}
 & FID $\downarrow$ & KID $\downarrow$ & FID$_{\text{dino}}$ $\downarrow$ & KID$_{\text{dino}}$ $\downarrow$ & CLIP $\uparrow$ & SigLIP2 $\uparrow$
 & CLIP $\uparrow$ & SigLIP2 $\uparrow$ \\
\midrule
\method{}, single-layer (layer 24)     & 25.21          & 0.101          & 0.121          & 0.0035          & 0.8977          & 0.9164          & 0.8081        & 0.8143 \\
\oursrow \method{}, multi-layer (layers 6, 12, 18, 24) & \textbf{20.99} & \textbf{0.033} & \textbf{0.100} & \textbf{0.0028} & \textbf{0.9062} & \textbf{0.9230} & \textbf{0.8519} & \textbf{0.8508} \\
\bottomrule
\end{tabular}
}
\end{table}

\section{Inference Details}
\label{app:inference_steps}

At test time, \method{} first generates a sparse voxel structure, then SLatFlow generates a PBR Gaussian latent at each active voxel (10 sampling steps), which decodes to PBR Gaussians or, optionally, a textured mesh.
The GS path has 4.5B parameters across sparse-structure flow, SLatFlow, and the shared SLatVAE decoder; the FlexiCubes mesh decoder brings the mesh path to 4.6B (Table~\ref{tab:generation}).

Table~\ref{tab:generation} reports the mean wall-clock time per generation over all inputs, \ie{}, the 412 Toys4K assets and 130 AI-generated images, on one NVIDIA H100 80GB GPU; for \method{}, this is ${\sim}$7\,s for the GS path and ${\sim}$67\,s (without tangent normal map) to ${\sim}$76\,s (with) for the mesh path. Times are measured warm: each stage first runs once on the same input in the same process, and that run is discarded. Timing stops at the output we evaluate: the textured mesh for mesh-based methods, the Gaussians for Gaussian-based methods, and, for LiTo, the Gaussians plus its untextured mesh. It excludes model loading, background removal, and rendering. TRELLIS~2's default low-VRAM mode, which \method{} also uses for its sparse-structure stage, moves weights between CPU and GPU during generation; this transfer time is included.

For each baseline, we use the sampling-step counts recommended with its public release, and LiTo is sampled with its official defaults. Table~\ref{tab:sampling_steps} lists the sparse-structure-stage and structured-latent-stage (SLat) step counts behind the times in Table~\ref{tab:generation}, including ours. LiTo outputs view-dependent 3D Gaussians; its auxiliary FlexiCubes head predicts geometry and vertex normals but not color, so its released pipeline does not output a textured mesh. Like the other Gaussian rows in Table~\ref{tab:generation}, LiTo is evaluated with image-space metrics, and the point-cloud metrics (ULIP, Uni3D-L) are omitted, since in our setup their point clouds are sampled from a textured mesh.

\begin{table}[h]
\centering
\caption{\textbf{Inference sampling steps per method.} SLat = structured-latent stage. ``---'' indicates a single-stage method without a separate structure flow.}
\label{tab:sampling_steps}
\begin{tabular}{l c c}
\toprule
Method & Sparse-structure stage & SLat \\
\midrule
3DTopia-XL~\citep{chen2024primx}        & ---  & 25 (DDIM) \\
TRELLIS~\citep{xiang2024trellis}        & 25   & 25 \\
LiTo~\citep{chang2026lito}              & ---  & 20 (Heun) \\
TRELLIS~2~\citep{xiang2025trellis2}     & 12   & 12 (shape) + 12 (texture) \\
\midrule
\oursrow \method{}~(ours)               & 12   & 10 \\
\bottomrule
\end{tabular}
\end{table}

\section{Evaluation Rendering and Illumination}
\label{app:eval_render}

All evaluation renders use CC0 HDRI environment maps from Poly Haven~\citep{polyhaven_hdris}. Generated assets are evaluated under the \emph{ninomaru\_teien} environment map on both the Toys4K and AI-generated-image benchmarks (Table~\ref{tab:generation}); the Toys4K ground truth is rendered under the same map. On Toys4K, the conditioning view is rendered under \emph{studio\_small\_01}; since the evaluation illumination (\emph{ninomaru\_teien}) differs from the conditioning illumination (\emph{studio\_small\_01}), methods that decompose materials (\method{}, TRELLIS~2, 3DTopia-XL) relight to the evaluation environment, while methods with baked appearance (TRELLIS~GS, LiTo) carry the conditioning illumination. On the AI-generated-image benchmark, the condition image carries no controllable illumination. The qualitative relighting figures also use Poly Haven maps: Fig.~\ref{fig:teaser} uses \emph{spruit\_sunrise}, \emph{studio\_small\_01}, and \emph{courtyard}; Fig.~\ref{fig:gen_compare_luce} uses \emph{studio\_small\_01}; Fig.~\ref{fig:gen_qual} and Fig.~\ref{fig:additional_qual} use \emph{spruit\_sunrise}; and Fig.~\ref{fig:vae_relight} cycles through six maps (\emph{venice\_sunset}, \emph{spruit\_sunrise}, \emph{moonless\_golf}, \emph{studio\_small\_01}, \emph{hotel\_room}, \emph{courtyard}).

\paragraph{Rendering.}
Meshes, including the Toys4K ground truth, are rendered in Blender~\citep{Blender}, and PBR Gaussians are shaded with split-sum image-based lighting (Eq.~\ref{eq:pbr_shading}). The Gaussians of TRELLIS~GS and LiTo have no material channels, so they cannot be relit and are rasterized as is. TRELLIS's mesh export bakes the Gaussian colors into a base-color texture with no metallic or roughness maps. As in 3DTopia-XL~\citep{chen2024primx}, which assigns a default diffuse material to baselines that cannot produce PBR materials, we render this mesh with its base-color texture and a constant non-metallic material. All evaluation renders are $1024\times1024$ on a white background. The camera follows the evaluation rig of TRELLIS~2~\citep{xiang2025trellis2}: the four yaws of Appendix~\ref{app:baseline_compare} at $30^{\circ}$ elevation, with a narrow $6^{\circ}$ FoV at a camera distance of 10.

\paragraph{Distributional metrics.}
FID~\citep{heusel2017gans} and KID~\citep{binkowski2018demystifying} compare renders of the generated assets with renders of the 412 ground-truth Toys4K assets. Each asset is rendered from these four views, the first at the viewpoint of the condition image. For each method, the color renders of all assets and views form one set, compared against the 1,648 ground-truth renders (412 assets $\times$ 4 views). Both KID variants are reported $\times 100$.

\section{Additional Qualitative Results}
\label{app:additional_qual}

\begin{figure}[h]
\begin{center}
\newcommand{\addqualcell}[1]{%
\begin{tikzpicture}[inner sep=0pt, outer sep=0pt]
  \node[anchor=north west] (strip) at (0,0)
       {\includegraphics[width=0.360\linewidth,trim=1024 0 2048 0,clip]{#1}};
  \node[anchor=east, draw=gray, line width=0.25pt, xshift=-2pt] at (strip.west)
       {\includegraphics[width=0.130\linewidth,trim=120 0 3806 0,clip]{#1}};
\end{tikzpicture}%
}
\newcommand{\addqualheader}{%
\makebox[0.130\linewidth][c]{\scriptsize Condition Image}\hspace{2pt}%
\makebox[0.180\linewidth][c]{\scriptsize PBR GS}%
\makebox[0.180\linewidth][c]{\scriptsize Render}%
}
\setlength{\tabcolsep}{1pt}
\renewcommand{\arraystretch}{0.9}
\begin{tabular}{@{}cc@{}}
\addqualheader & \addqualheader \\
\addqualcell{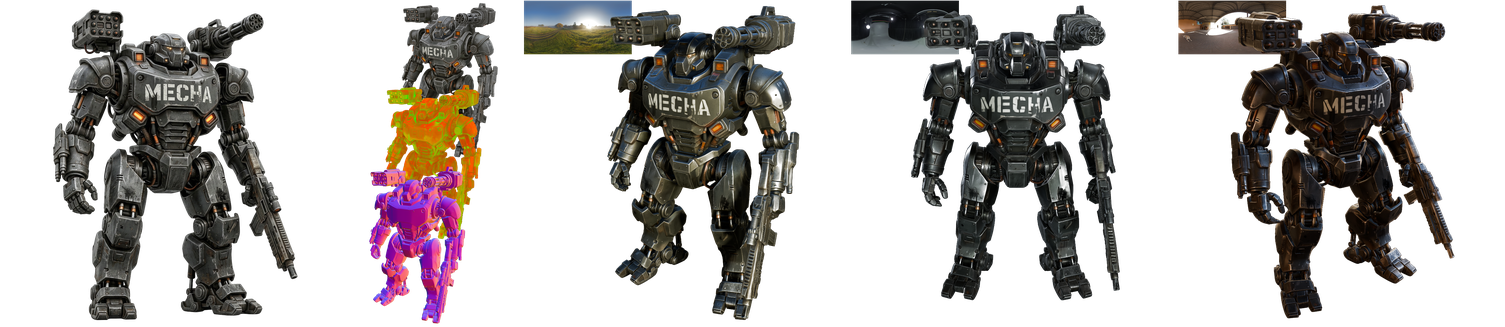} &
\addqualcell{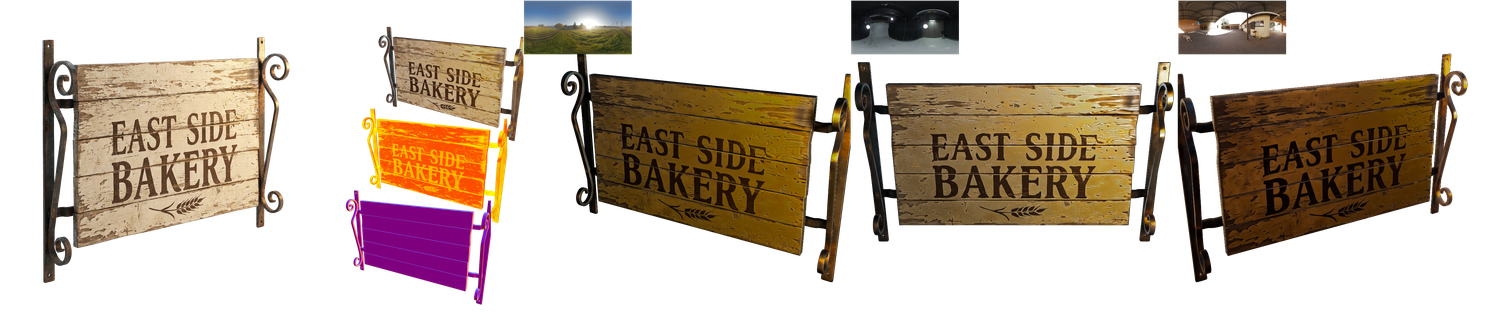} \\
\addqualcell{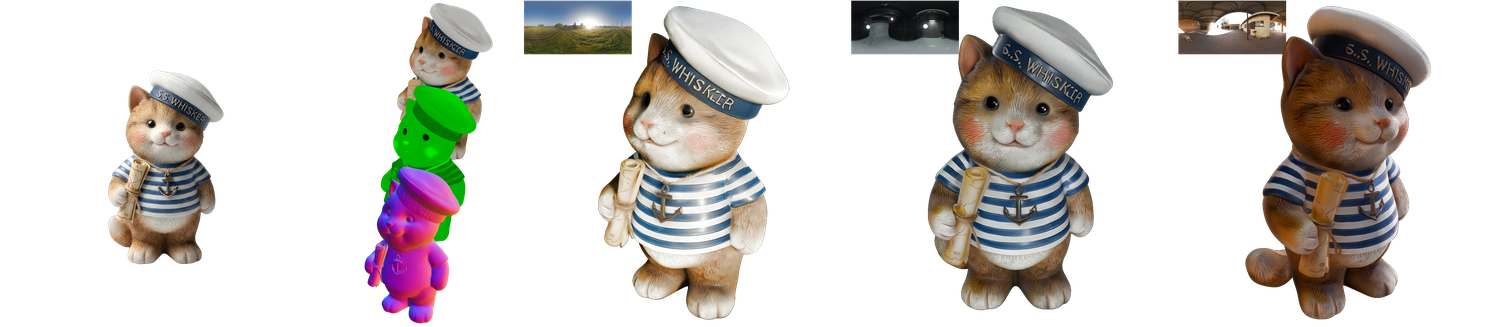} &
\addqualcell{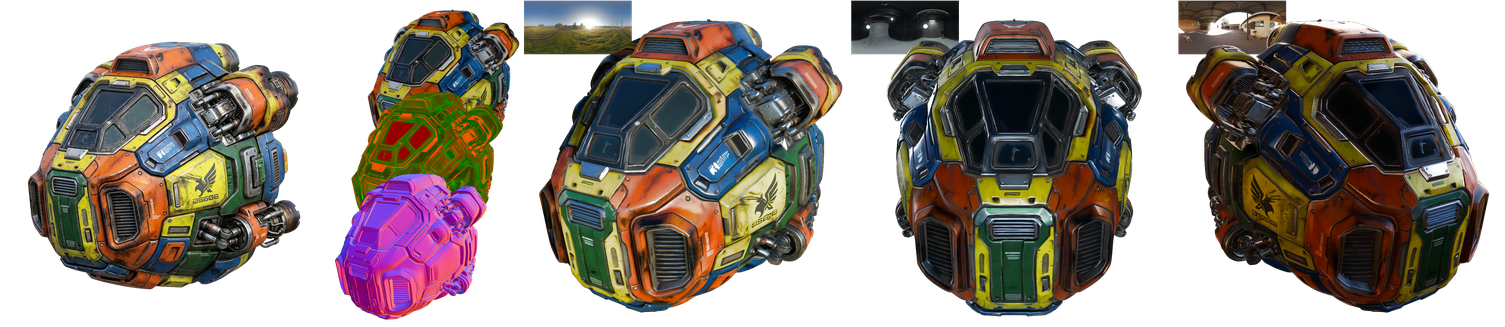} \\
\addqualcell{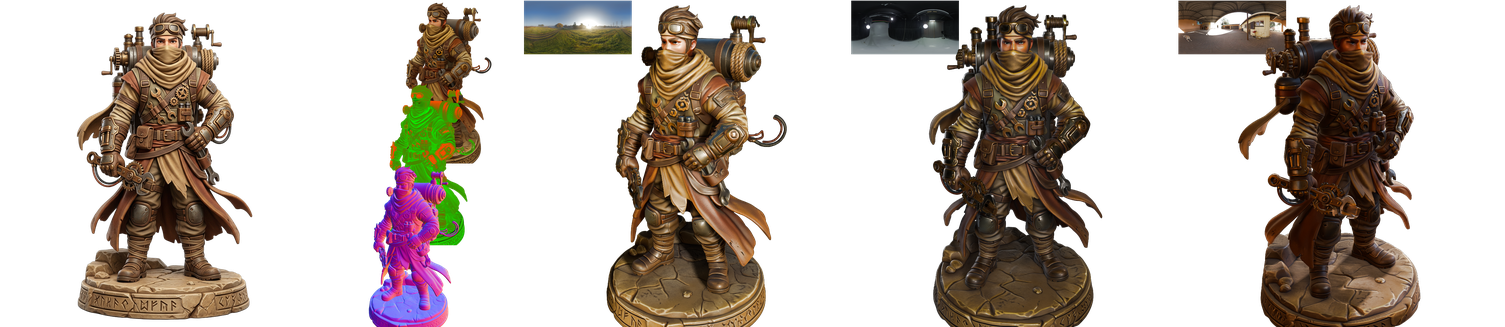} &
\addqualcell{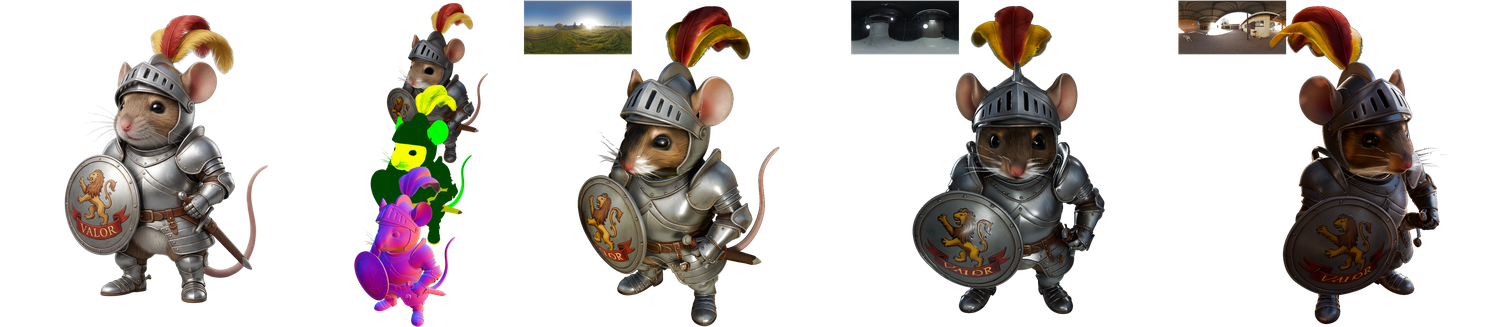} \\
\addqualcell{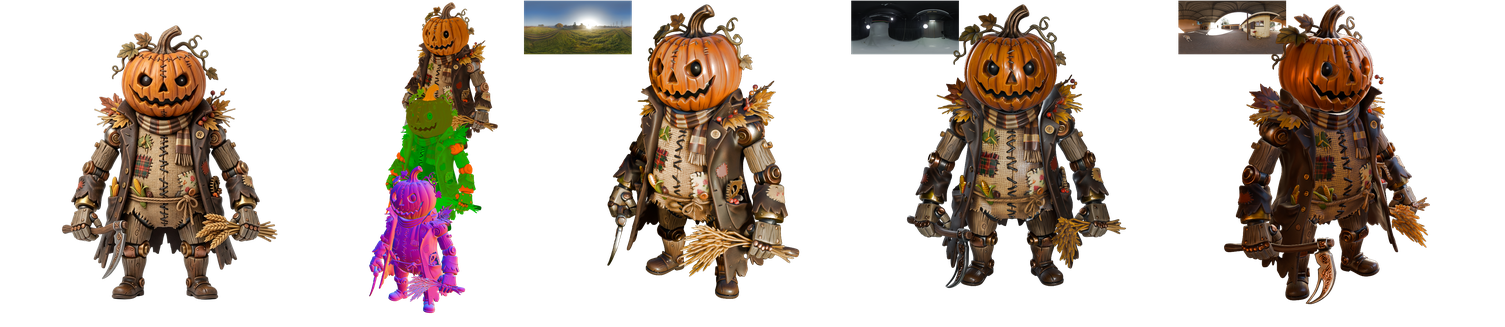} &
\addqualcell{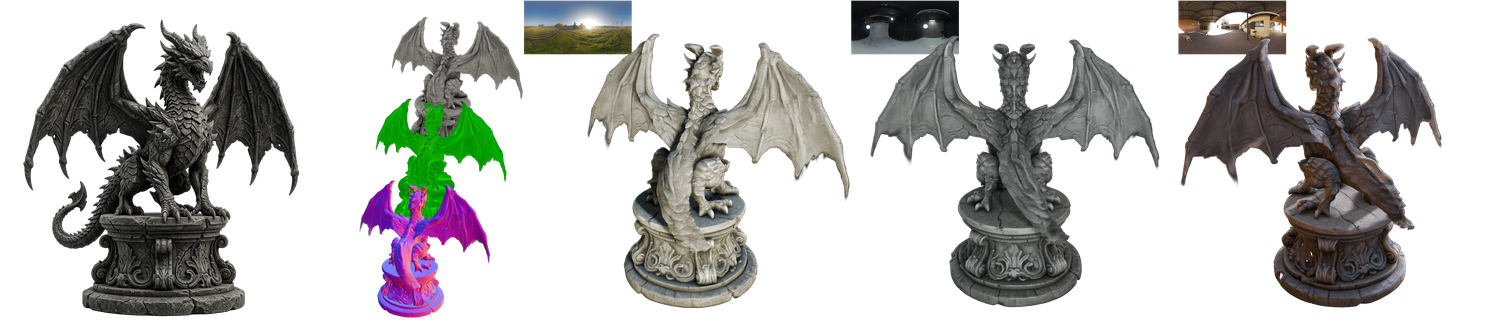} \\
\addqualcell{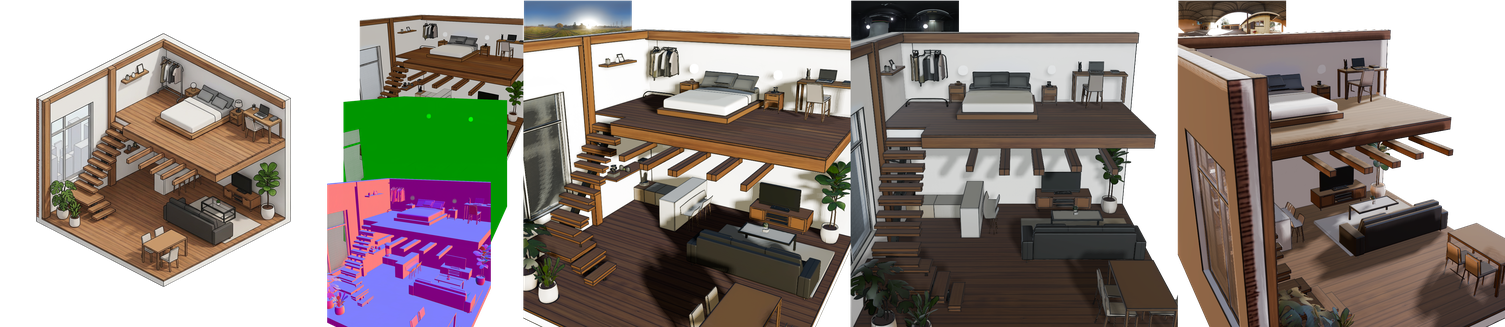} &
\addqualcell{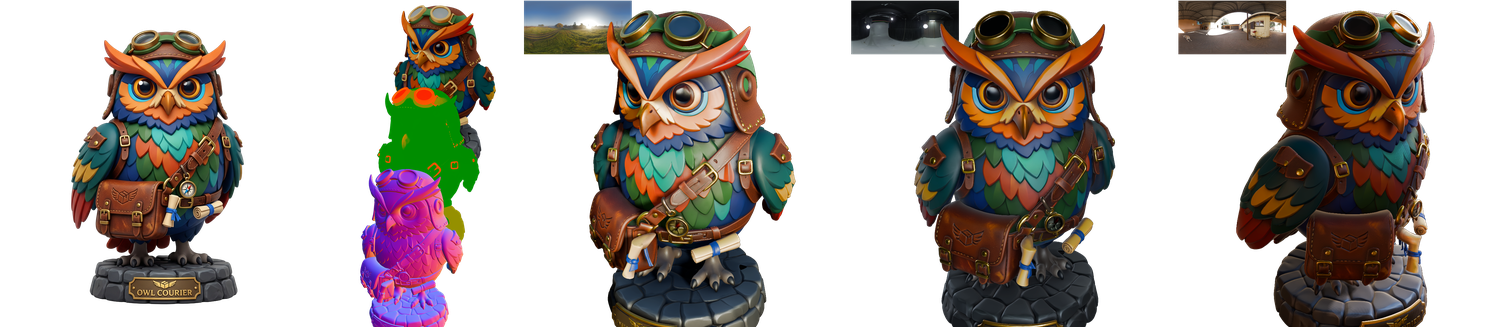} \\
\end{tabular}
\end{center}
\vspace{-0.3cm}
\caption{\textbf{Additional generation results.}
Additional \method{} examples across diverse asset categories. Each cell shows, from left to right, the condition image, the per-modality PBR Gaussians (albedo, metallic-roughness, normals), and a shaded render under the environment map previewed above it.}
\label{fig:additional_qual}
\end{figure}
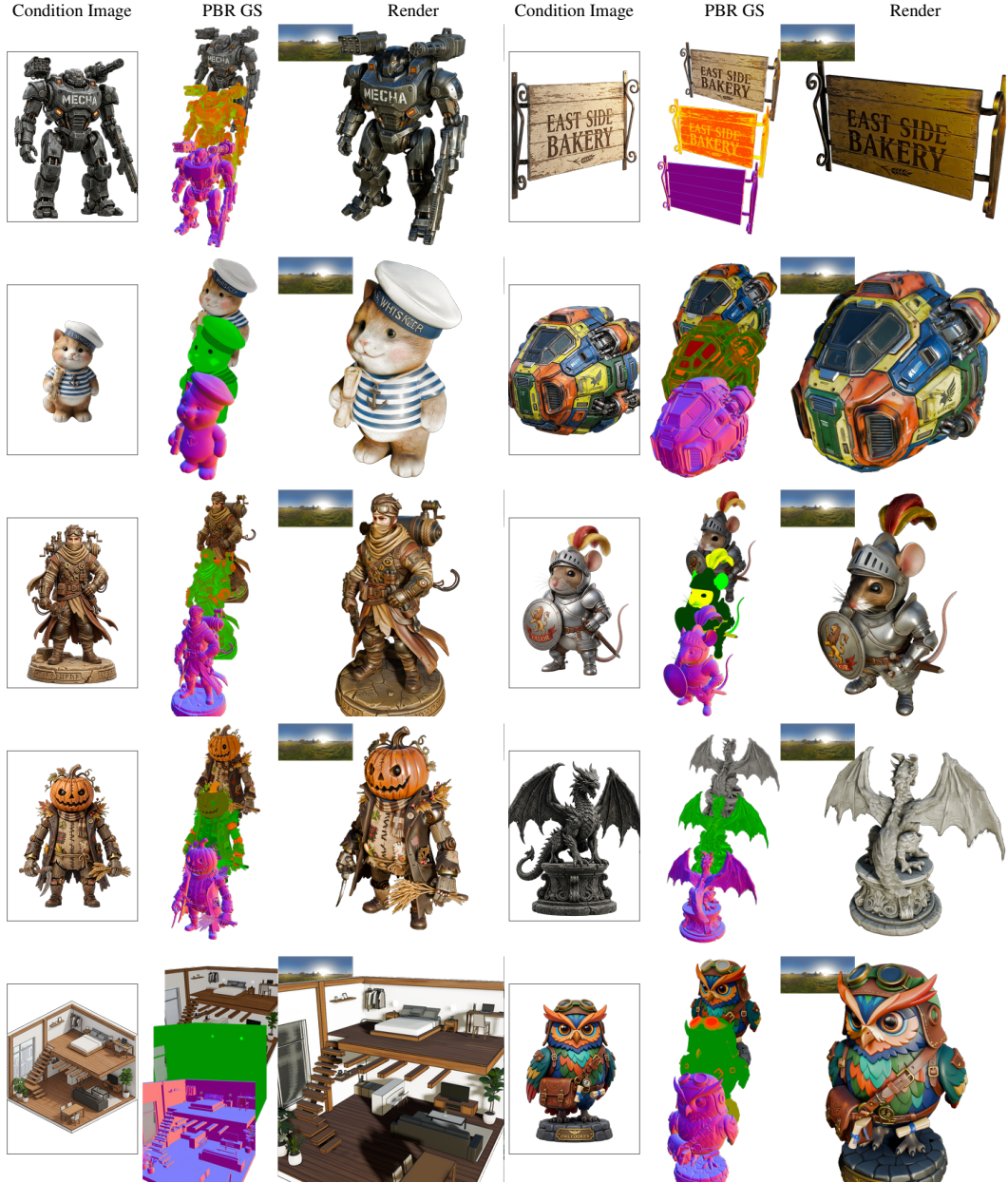

Figure~\ref{fig:additional_qual} shows additional generation results across diverse asset categories, including furniture, vehicles, characters, and household objects.
Each result shows the input condition image, the generated asset's per-modality PBR Gaussians (albedo, metallic-roughness, normals), and a shaded render.

\section{Baseline Comparisons on the AI-Generated-Image Benchmark}
\label{app:baseline_compare}

\begin{figure}[p]
\centering
\includegraphics[height=0.84\textheight, keepaspectratio]{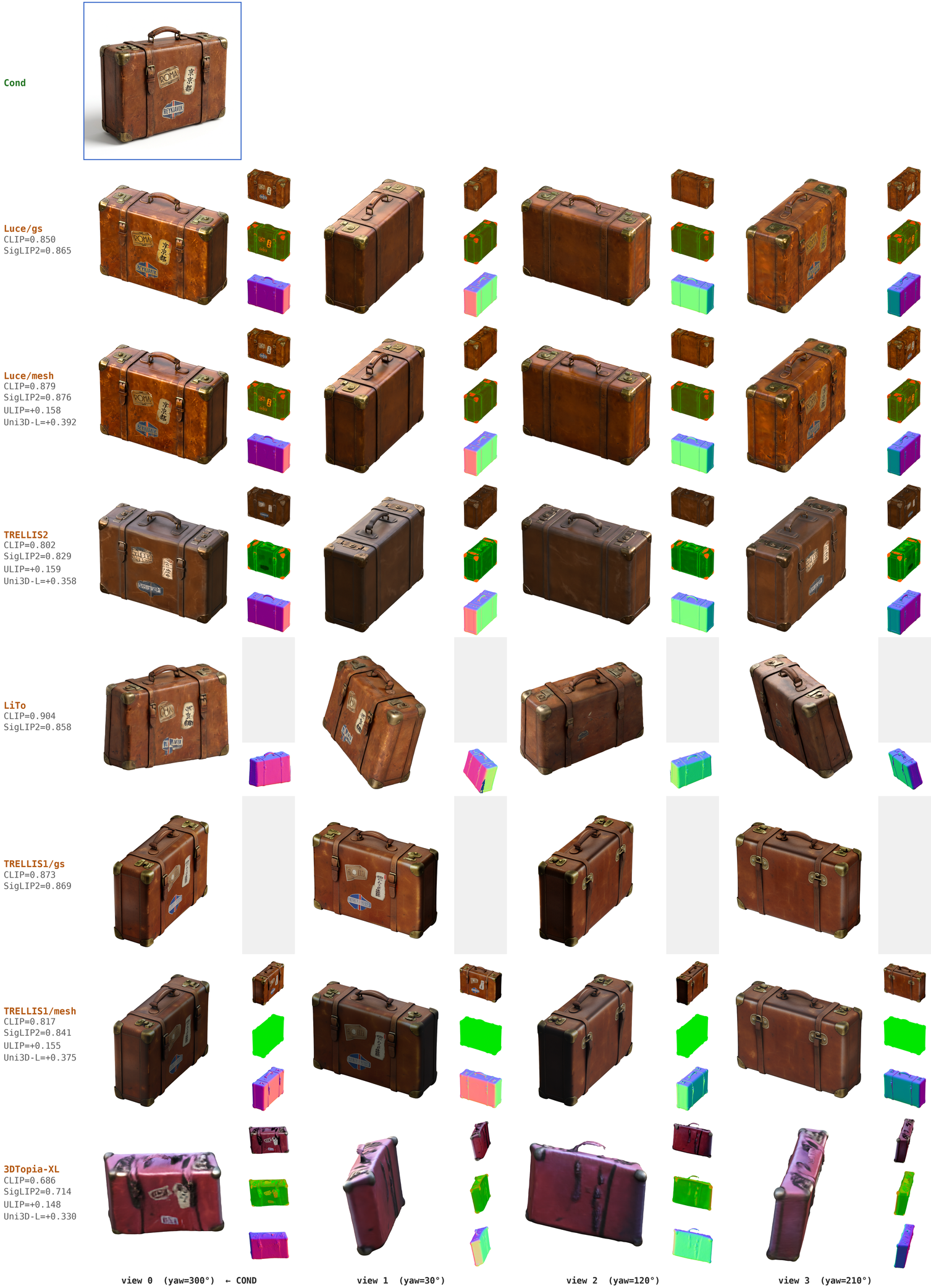}
\caption{\textbf{Per-sample baseline comparison (1/8).} Rows: methods (top label = condition); \method{} rows are ours. Columns: four evaluation views (yaws $300^{\circ}$, $30^{\circ}$, $120^{\circ}$, $210^{\circ}$) matching the views used to compute the CLIP and SigLIP2 scores in Table~\ref{tab:generation}. Per-row metrics show CLIP and SigLIP2, plus ULIP and Uni3D-L for the mesh rows, on the depicted sample. For methods with explicit material decomposition, three small thumbnails are stacked to the right of each shaded render, top to bottom: albedo, metallic-roughness (metallic=R, roughness=G, B unused), and surface normals. LiTo has only the normal thumbnail and TRELLIS~GS none; TRELLIS~mesh has only a base-color texture, so its uniform metallic-roughness thumbnail is the default material it is rendered with (Appendix~\ref{app:eval_render}).}
\label{fig:baseline_compare_1}
\end{figure}

\begin{figure}[p]
\centering
\includegraphics[width=\textwidth, height=0.92\textheight, keepaspectratio]{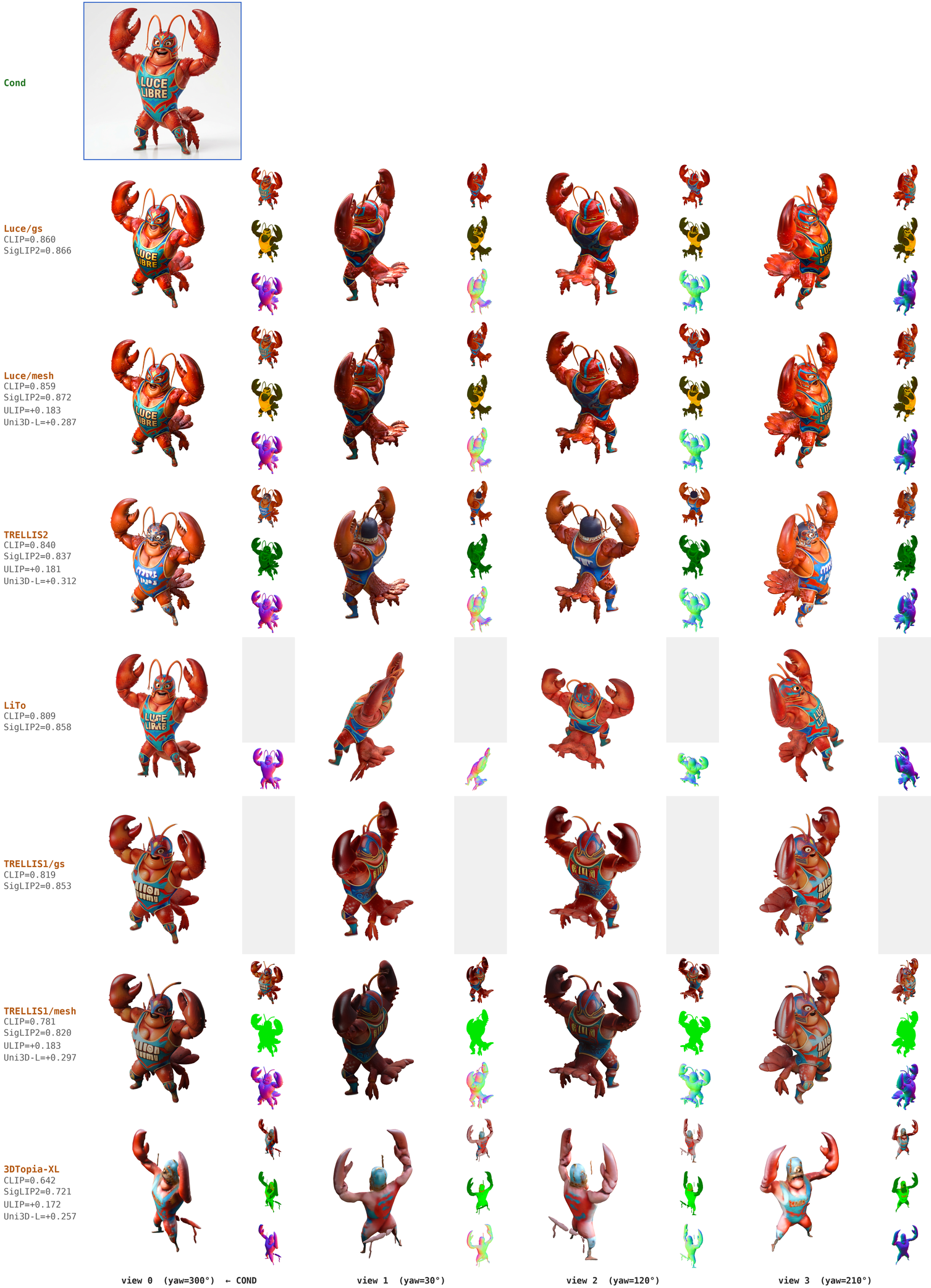}
\caption{\textbf{Per-sample baseline comparison (2/8).} Same layout as Figure~\ref{fig:baseline_compare_1}.}
\label{fig:baseline_compare_2}
\end{figure}

\begin{figure}[p]
\centering
\includegraphics[width=\textwidth, height=0.92\textheight, keepaspectratio]{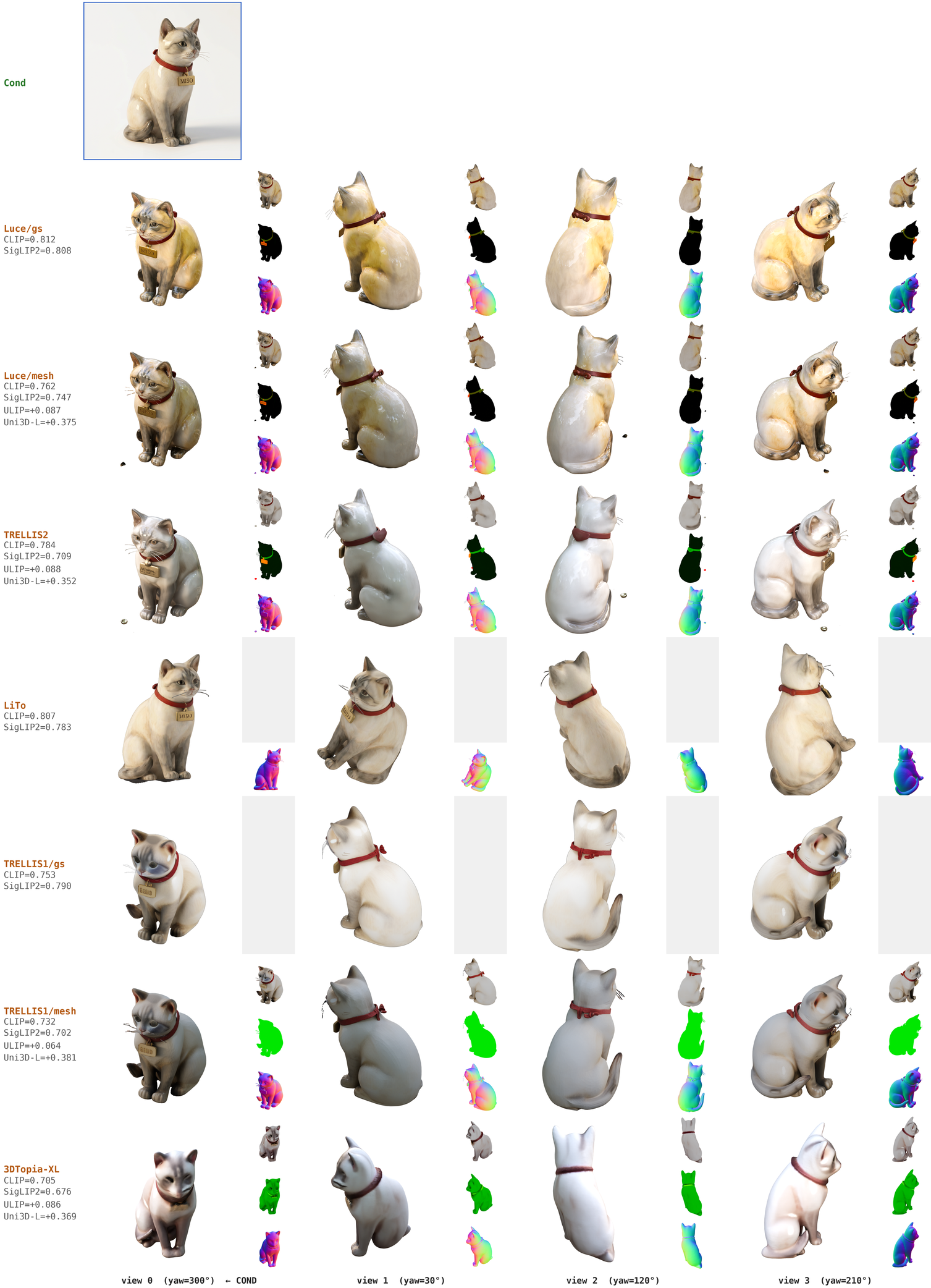}
\caption{\textbf{Per-sample baseline comparison (3/8).} Same layout as Figure~\ref{fig:baseline_compare_1}.}
\label{fig:baseline_compare_3}
\end{figure}

\begin{figure}[p]
\centering
\includegraphics[width=\textwidth, height=0.92\textheight, keepaspectratio]{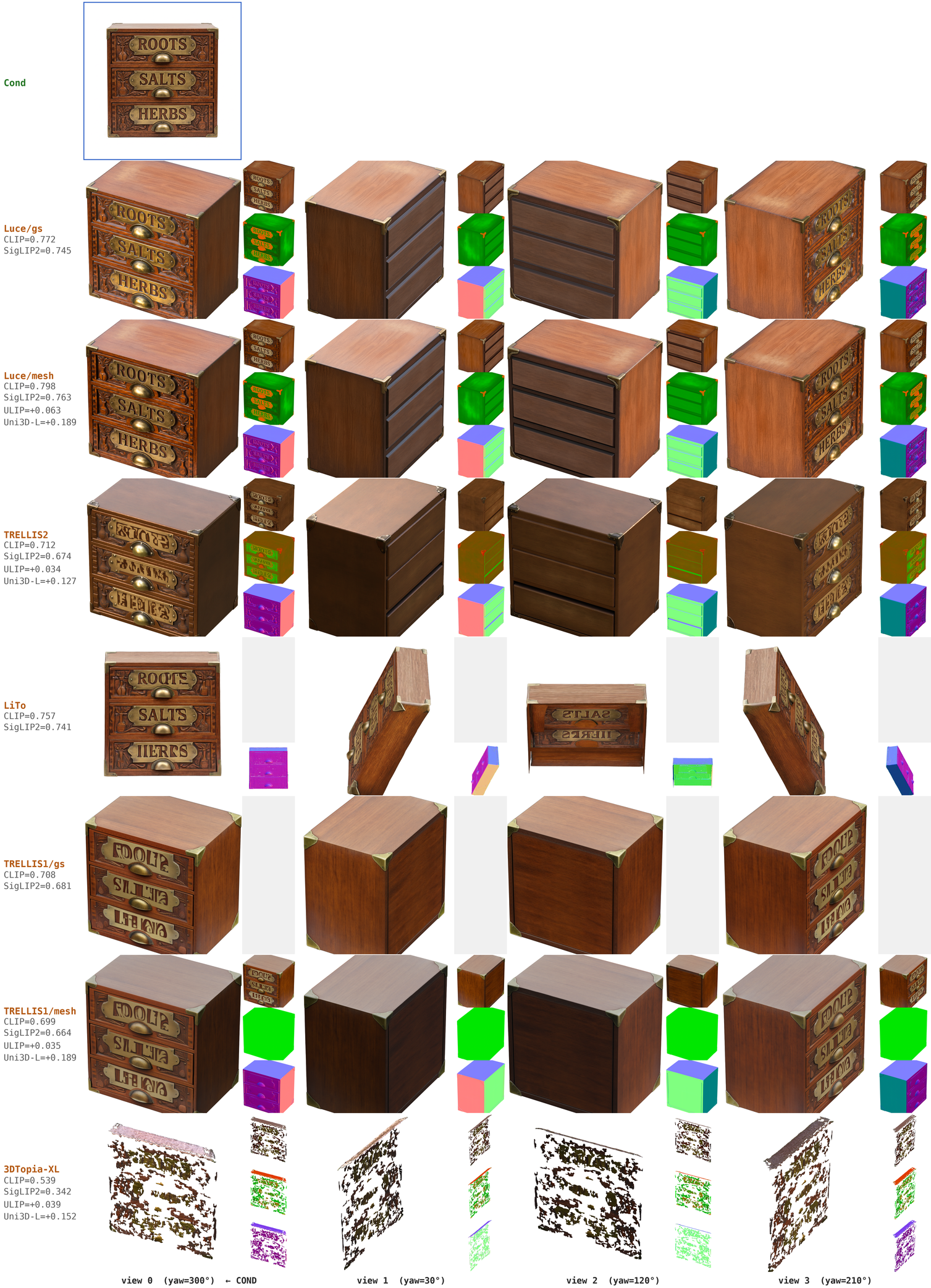}
\caption{\textbf{Per-sample baseline comparison (4/8).} Same layout as Figure~\ref{fig:baseline_compare_1}.}
\label{fig:baseline_compare_4}
\end{figure}

\begin{figure}[p]
\centering
\includegraphics[width=\textwidth, height=0.92\textheight, keepaspectratio]{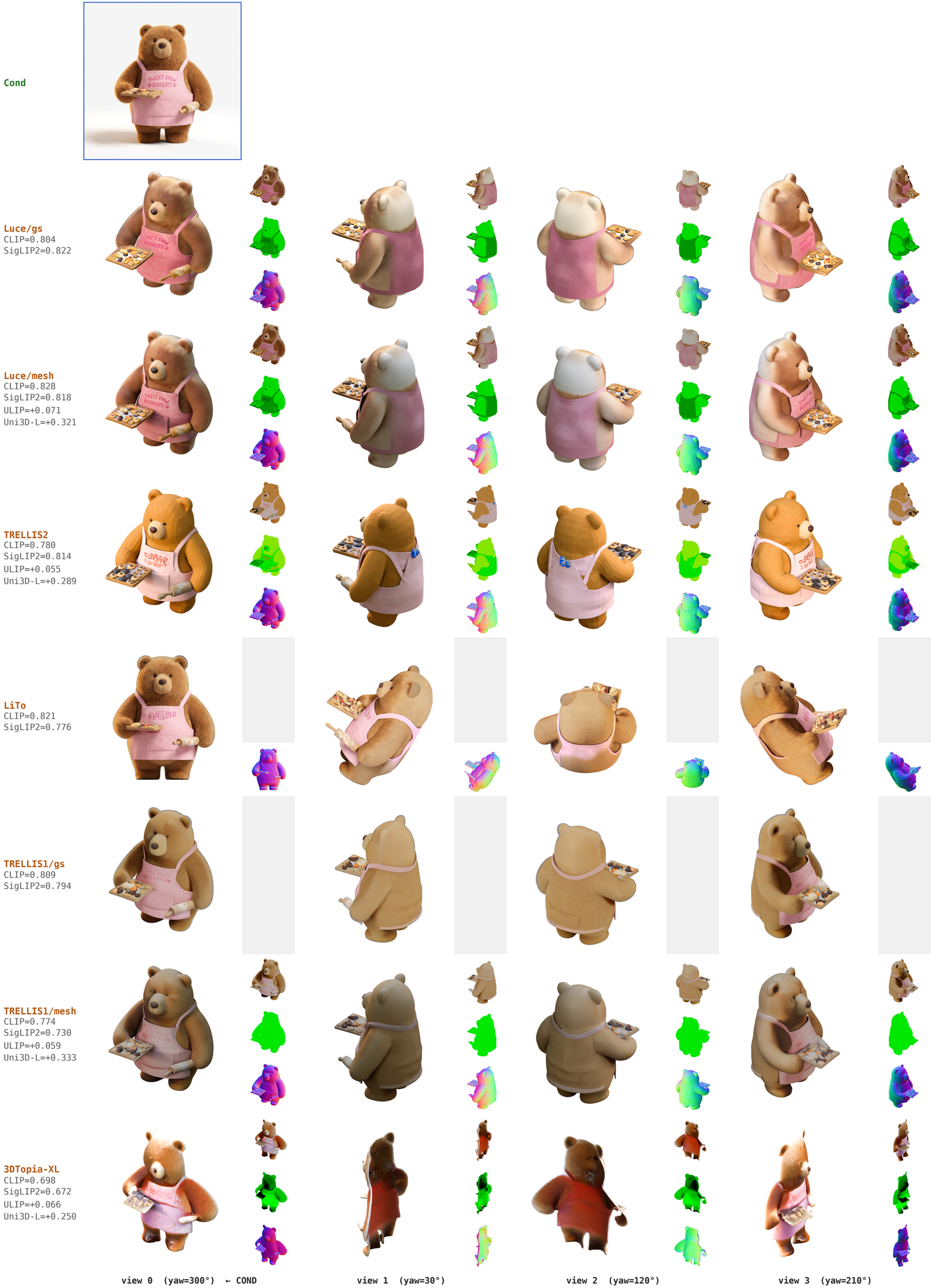}
\caption{\textbf{Per-sample baseline comparison (5/8).} Same layout as Figure~\ref{fig:baseline_compare_1}.}
\label{fig:baseline_compare_5}
\end{figure}

\begin{figure}[p]
\centering
\includegraphics[width=\textwidth, height=0.92\textheight, keepaspectratio]{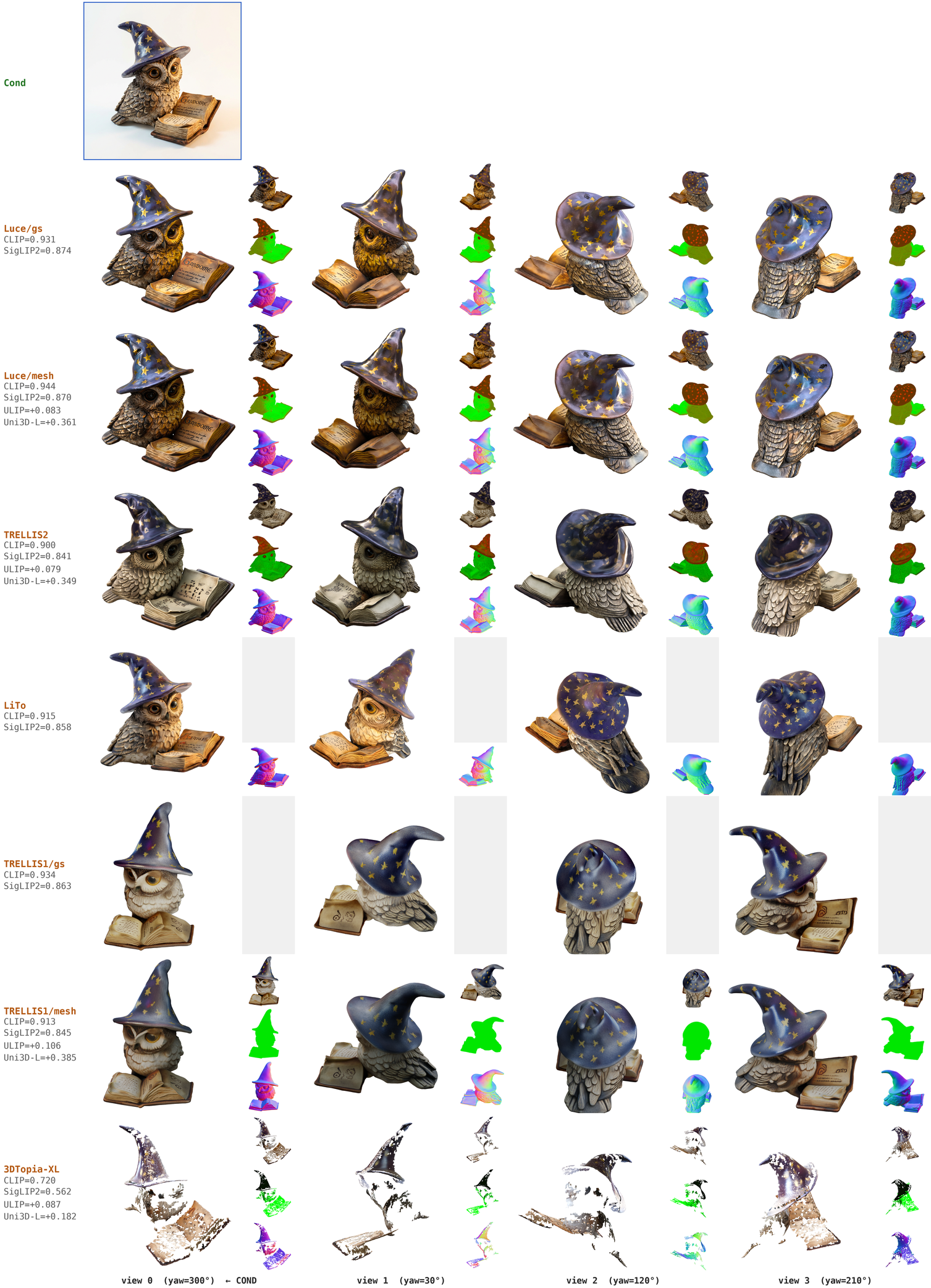}
\caption{\textbf{Per-sample baseline comparison (6/8).} Same layout as Figure~\ref{fig:baseline_compare_1}.}
\label{fig:baseline_compare_6}
\end{figure}

\begin{figure}[p]
\centering
\includegraphics[width=\textwidth, height=0.92\textheight, keepaspectratio]{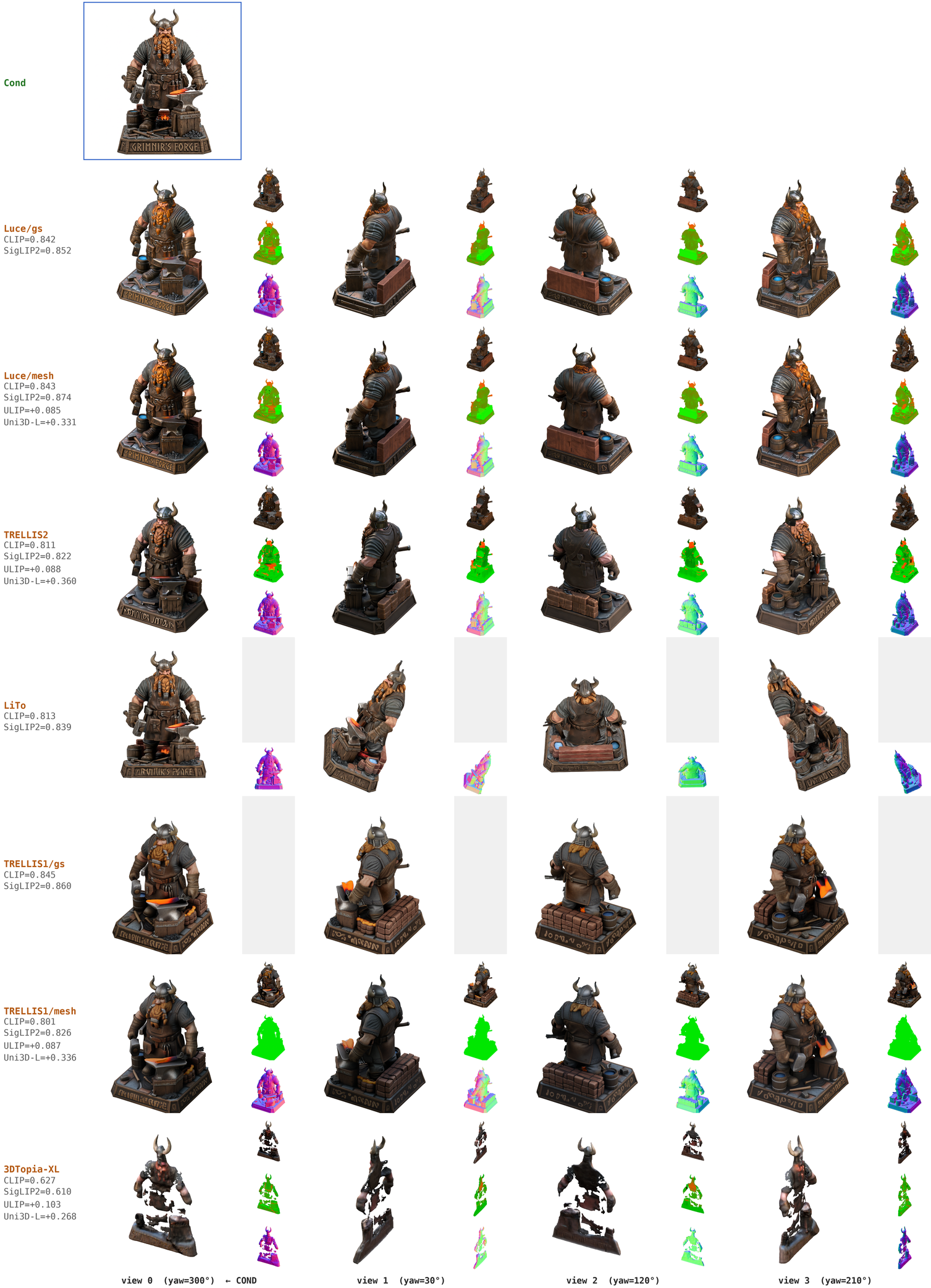}
\caption{\textbf{Per-sample baseline comparison (7/8).} Same layout as Figure~\ref{fig:baseline_compare_1}.}
\label{fig:baseline_compare_9}
\end{figure}

\begin{figure}[p]
\centering
\includegraphics[width=\textwidth, height=0.92\textheight, keepaspectratio]{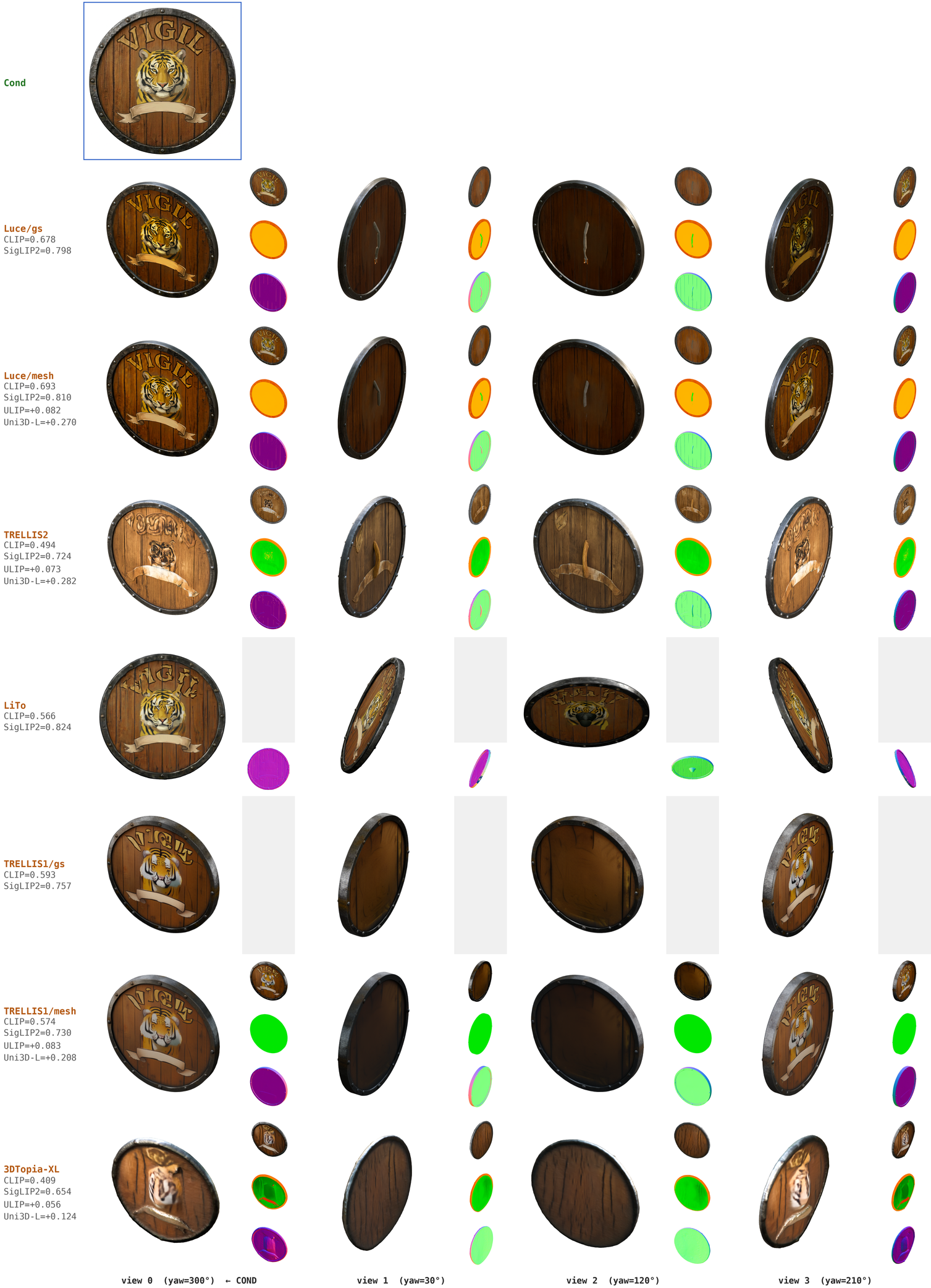}
\caption{\textbf{Per-sample baseline comparison (8/8).} Same layout as Figure~\ref{fig:baseline_compare_1}.}
\label{fig:baseline_compare_10}
\end{figure}

To complement the quantitative results of Table~\ref{tab:generation} and the qualitative summary in Fig.~\ref{fig:gen_compare}, we include per-sample comparisons against every baseline on eight representative inputs from our AI-generated-image benchmark. Each mosaic (Figures~\ref{fig:baseline_compare_1}--\ref{fig:baseline_compare_10}) follows the same layout: rows are methods, columns are the four evaluation views (yaws $300^{\circ}$, $30^{\circ}$, $120^{\circ}$, $210^{\circ}$, evenly spaced at $90^{\circ}$ around the asset at $30^{\circ}$ elevation), with the condition image shown in the top-left. Camera and lighting follow Appendix~\ref{app:eval_render}. \textbf{These four views are the exact rendered views over which we average CLIP and SigLIP2 in Table~\ref{tab:generation}}; ULIP and Uni3D-L, reported for mesh rows only, are computed on point clouds sampled from the textured mesh (Appendix~\ref{app:inference_steps}). The per-sample scores in each row label follow the same protocol, so they are directly comparable to the dataset-level averages in Table~\ref{tab:generation}. Because LiTo generates assets in the input view's frame rather than a canonical orientation (Fig.~\ref{fig:gen_compare_luce}), the same four yaws place its views closer to the conditioning viewpoint than the other methods', so its view set is not pose-matched to theirs. The \method{} mesh row uses the tangent-space normal map; Table~\ref{tab:generation} also reports the variant without it.
The selected samples span a range of input difficulty: heavy texture detail, fine-grained geometry, mixed-material surfaces, and assets containing legible text or logos. Across the eight mosaics, \method{} generally produces sharper material decomposition, more detailed normal maps, and better preservation of fine spatial detail than the closest baseline.

\clearpage

\clearpage

\clearpage

\clearpage

\clearpage

\clearpage

\clearpage

\clearpage

\end{document}